\documentclass[10pt,twocolumn,letterpaper]{article}
\usepackage{titling}
\usepackage{iccv}
\usepackage{times}
\usepackage{epsfig}
\usepackage{graphicx}
\usepackage{amsmath}
\usepackage{amssymb}
\usepackage{adjustbox}
\usepackage{multirow}
\usepackage{caption}
\usepackage[accsupp]{axessibility}
\usepackage[colorlinks,linkcolor=blue]{hyperref}

\iccvfinalcopy 

\def\iccvPaperID{****} 

\ificcvfinal\fi

\begin{document}

\title{Self-supervised Learning of Implicit Shape Representation with\\
Dense Correspondence for Deformable Objects}

\author{
Baowen Zhang$^{1,2}$\qquad Jiahe Li$^{1,2}$\qquad Xiaoming Deng$^{1,2}$\thanks{indicates corresponding author}\qquad Yinda Zhang$^3$$^*$ \\   Cuixia Ma$^{1,2}$\qquad Hongan Wang$^{1,2}$\\
$^1$Institute of Software, Chinese Academy of Sciences \quad $^2$University of Chinese Academy of Sciences\\  $^3$Google \quad  
}

\maketitle
\ificcvfinal\thispagestyle{empty}\fi

\begin{abstract}
Learning 3D shape representation with dense correspondence for deformable objects is a fundamental problem in computer vision. 
Existing approaches often need additional annotations of specific semantic domain, e.g\onedot, skeleton poses for human bodies or animals, which require extra annotation effort and suffer from error accumulation, and they are limited to specific domain.
In this paper, we propose a novel self-supervised approach to learn neural implicit shape representation for deformable objects, which can represent shapes with a template shape and dense correspondence in 3D.
Our method does not require the priors of skeleton and skinning weight, and only requires a collection of shapes represented in signed distance fields.
To handle the large deformation, 
we constrain the learned template shape in the same latent space with the training shapes,
design a new formulation of local rigid constraint that enforces rigid transformation in local region and addresses local reflection issue, and present a new hierarchical rigid constraint to reduce the ambiguity due to the joint learning of template shape and correspondences.
Extensive experiments show that our model can represent shapes with large deformations. 
We also show that our shape representation can support two typical applications, such as texture transfer and shape editing, with competitive performance.
The code and models are available at \href{https://iscas3dv.github.io/deformshape}{https://iscas3dv.github.io/deformshape}.
\end{abstract}

\section{Introduction}
    \begin{figure}
    \centering
    \includegraphics[width=1.0\linewidth]{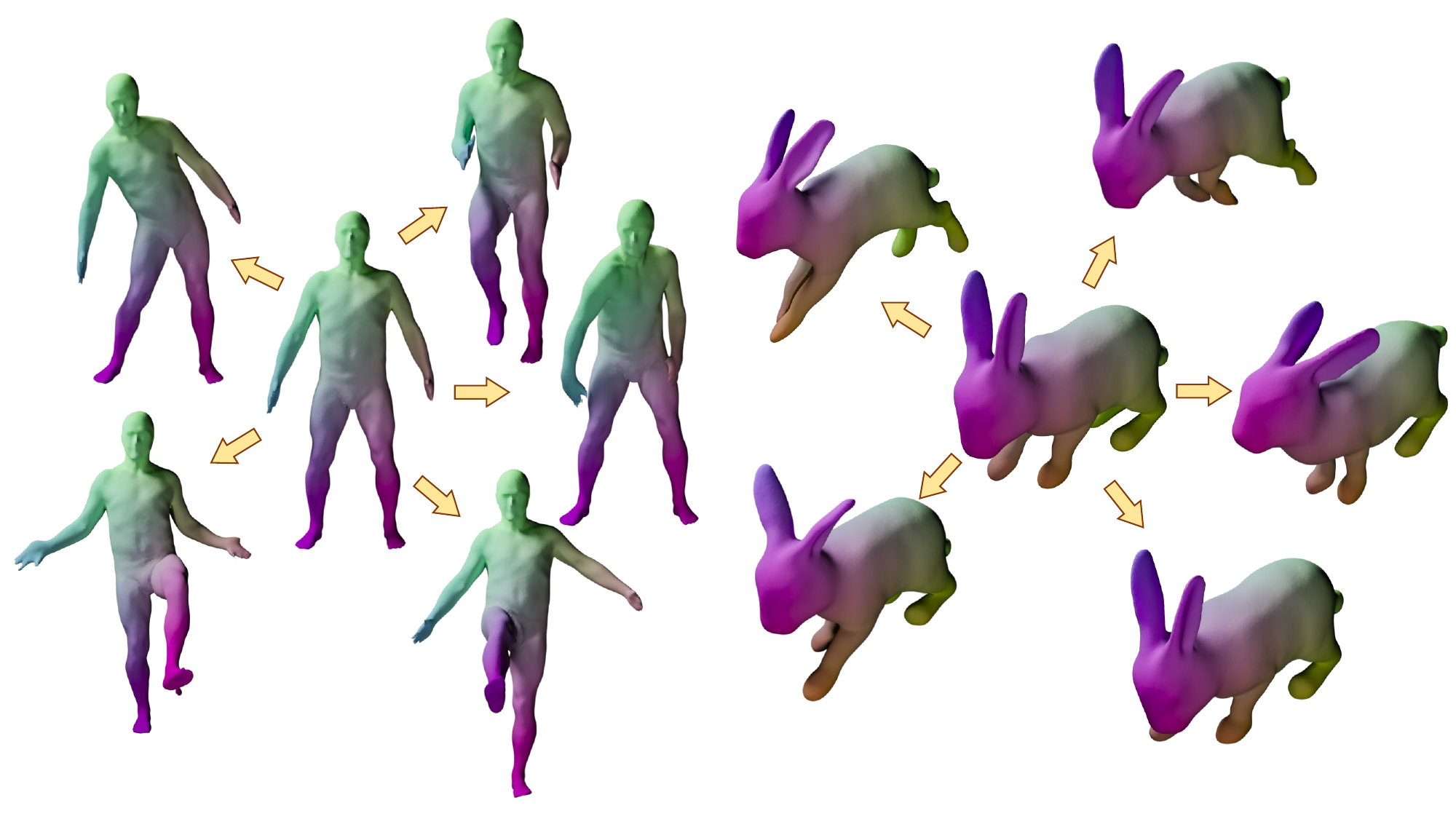}
    \caption{We present a self-supervised method to learn neural
    implicit representation for deformable objects with a collection of shapes. 
    Our method can generate shapes by deforming a learned template and get dense correspondence.
    }
    \label{fig:teaser}
\end{figure}
Shape representation with dense correspondence is a fundamental problem in computer vision. It plays a key role in many applications such as shape reconstruction~\cite{newcombe2015dynamicfusion,saito2019pifu, SkiRT:3DV:2022}, texture mapping~\cite{raj2021anr,Deng_2021_CVPR}, and shape editing~\cite{Deng_2021_CVPR, deng2020nasa, sorkine2007rigid}.
Early works often need additional semantic prior or annotations to learn such representation, e.g., SMPL~\cite{loper2015smpl} and SMAL~\cite{Zuffi:CVPR:2017} use registered meshes in certain semantic categories and LEAP ~\cite{Mihajlovic_2021_CVPR} and NASA~\cite{deng2020nasa} require annotations of skeletons and skinning weights, which limits the use cases and scalability of these representations.
On the other hand, with the emerging implicit representation, more 3D assets are encoded in implicit sign distance function (SDF), and a shape representation for deformable objects in SDF is still largely missing in the community.

In this paper, we aim to design a neural representation for deformable objects (Figure~\ref{fig:teaser}). 
Given a target deformable object represented as a set of sign distance field under various deformations, our method learns an implicit representation that is able to reconstruct the 3D shapes, interpolate between the given examples, and provide dense correspondence across shapes, in a fully self-supervised manner without any additional annotation or semantic prior.
Following the common idea \cite{Deng_2021_CVPR,park2021nerfies}, we formulate the deformable shape as a static shape in a canonical (or template) space, plus the mappings from any target deformation space to the template space for arbitrary locations in the 3D space.
However, in the existence of large deformation, such as humans and animals, we empirically found the above-mentioned approaches tend to be unstable and can easily get stuck in local optima if learned self-supervised due to unique challenges:

\vspace{1mm}
\noindent\textbf{Highly Under-constrained Optimization.}
The free-form template shape and deformation need optimizing jointly but highly under-constrained.
Under the case with large deformation across training examples, the per-location mapping is error-prone, which in return affects the template shape, and there lacks a good constraint to push both back to the correct case.
To mitigate the issue, we learn a generative model for the training shapes, where each shape is represented as a code in a latent space.
We then enforce a valid shape for the template shape by sampling from this latent space.
We found this helps constrain the template shape and benefits the learning of dense correspondence.

\vspace{1mm}
\noindent\textbf{Incomplete Local Rigid Constraint.}
As-rigid-as-possible (ARAP) constraint \cite{alexa2000rigid,sorkine2007rigid} has been extensively used for discretized surface such as meshes to penalize irregular surface deformation in many previous work \cite{alexa2000rigid,sorkine2007rigid,igarashi2005rigid}.
However, defining ARAP equivalent constraint on continuous SDF is non-trivial.
Existing works conduct a few attempts but all have their drawbacks.
For example, Deng \etal \cite{Deng_2021_CVPR} use smooth constraints to avoid predicted deformation being large. However, this work cannot model the shapes with large deformation. 
Park \etal \cite{park2021nerfies} propose a local rigid constraint however does not penalize flip mapping in a local region, and as a result a point on the left hand might be incorrectly mapped to the right hand.
In contrast, we propose a novel formulation of local rigid constraint that enforces rigid transformation in local region and addresses local flip mapping issue, as illustrated in reflection issue of Figure~\ref{fig:rigid}(a).
We give theoretical analysis and show that our local rigid constraint is, to the best of our knowledge, the first ARAP equivalent constraint define with implicit representation in infinitesimal scopes.

\vspace{1mm}
\noindent\textbf{Insufficient Large-scale Deformation Prior.}
We found the learned shape representations with only local rigid constraint at each point \cite{park2021nerfies,Deng_2021_CVPR} still suffer from local optima.
This is because shapes with large deformation often have rigid deformation in large scope, such as small neighborhood rigid regions (Figure~\ref{fig:rigid}(b)) and large rigid regions such as the limb on human body (Figure~\ref{fig:rigid}(c)), 
and the local rigid constraint does not have enough spatial context to take effect. 
Several rigid constraints on meshes \cite{zuffi2015stitched} or point clouds \cite{li2019lbs} have been designed and proved to be effective for shape representation by utilizing connections between surface points. 
However, it is not straight-forward to extend these rigid constraints to implicit shape representation due to the lack of explicit connections between points.
Therefore, previous implicit representation methods neglect large-scale deformation prior.
To this end, we design hierarchical rigid constraints for implicit representation to utilize  spatial context of shapes at rigid part level and neighborhood level to reduce the ambiguity to learn shape and correspondence, which effectively constrains rigid motion in large scale and stabilizes the learning of the representation.

We perform extensive experiments to verify that our neural representation, learned with three above mentioned contributions, has superior capability in shape reconstruction, deformation interpolation, and building dense correspondence.
We also show that high quality results can be achieved in various applications, including texture transfer and shape editing, using our learned representation.

\section{Related Work}

\vspace{1mm}
\noindent \textbf{Neural Implicit Representation for Rigid Object.}
Implicit function is widely used in 3D shape representation.
Park \etal \cite{park2019deepsdf} propose an efficient model named DeepSDF to learn SDF to represent shapes. Mescheder \etal \cite{mescheder2019occupancy} and Chen \etal \cite{chen2019learning} achieve neural implicit representation assignment by means of a binary classifier. 
Chibane \etal \cite{chibane2020neural} use unsigned distance field to achieve high resolution output of arbitrary shape.
Deng \etal \cite{Deng_2021_CVPR} propose DIF, which learns a template to deform to a class of objects. 
This method works well for rigid objects, but fails when large deformation occurs, such as moving human body.

\vspace{1mm}
\noindent \textbf{Neural Shape Representation with Shape Priors.}
Several methods present neural shape representation for dynamic shapes using shape priors such as skeleton and skinning for human body.  
Prior art works on neural representation of human body \cite{deng2020nasa,Mihajlovic_2021_CVPR,chen2021snarf} impose bone transformation to constrain the deformation space. Jiang \etal \cite{jiang2021learning} use pose-shape pairs in training data to learn a model to represent moving body. Ma \etal \cite{Ma_2021_ICCV} project human body on a pre-defined UV map and represent body shape as point cloud. These methods require the pre-defined topology space, and the known or easy-to-learn diffeomorphism to realize human body shape representation. However, none of them can optimize template and correspondence simultaneously. 
Skeleton provides much prior of deformation, so methods using skeleton are advantageous when ground truth pose are available. In this paper, we focus on deformable shapes without pre-defined skeleton and propose a skeleton-free shape representation with dense correspondence.

\vspace{1mm}
\noindent \textbf{Neural Implicit Representation with Dense Correspondence.}
Training a neural implicit model with dense correspondence is a longstanding task \cite{Deng_2021_CVPR}. 
Oflow \cite{Oflow} can model motion sequence of a deformed shape. 
This method can model shapes with large deformation, but it requires continuous shape sequence as input. DIF \cite{Deng_2021_CVPR} can represent shapes of the same category and generate dense correspondences among shapes. However, DIF cannot model shapes with large deformation, such as human body.

\vspace{1mm}
\noindent \textbf{Neural Dense Correspondence.}
Finding accurate dense correspondence among shapes is a fundamental problem in computer vision. Several methods have explored this problem in supervised manner \cite{litany2017fmnet, groueix2018b, sundararaman2022implicit} or self-supervised manner \cite{groueix2018b, halimi2019unsupervised, bhatnagar2020loopreg, eisenberger2020deep, eisenberger2021neuromorph}. In supervised methods, the corresponding points between the input shapes are required to know. 
In self-supervised case, some works \cite{groueix2018b, halimi2019unsupervised, eisenberger2020deep, eisenberger2021neuromorph} use topology information to constrain the point location on the surface.
However, these self-supervised methods require mesh as input, which is harder to access than point cloud in real world capture. Several works in human mesh registration \cite{bhatnagar2020loopreg, trappolini2021shape} also predict dense correspondences between shapes. Bharat \etal \cite{bhatnagar2020loopreg} use the shape model SMPL \cite{loper2015smpl} as a prior for training, yet it cannot be generalized to other shape categories without a pre-built shape model. Giovanni \etal \cite{trappolini2021shape} present a learning approach to register
non-rigid 3D point clouds. However, this work requires ground truth correspondences for training. 
In this paper, we focus on learning dense correspondence with neural implicit representation using self-supervised method.

\section{Method}
    \begin{figure}[t]
    \centering
    \includegraphics[width=0.5\textwidth]{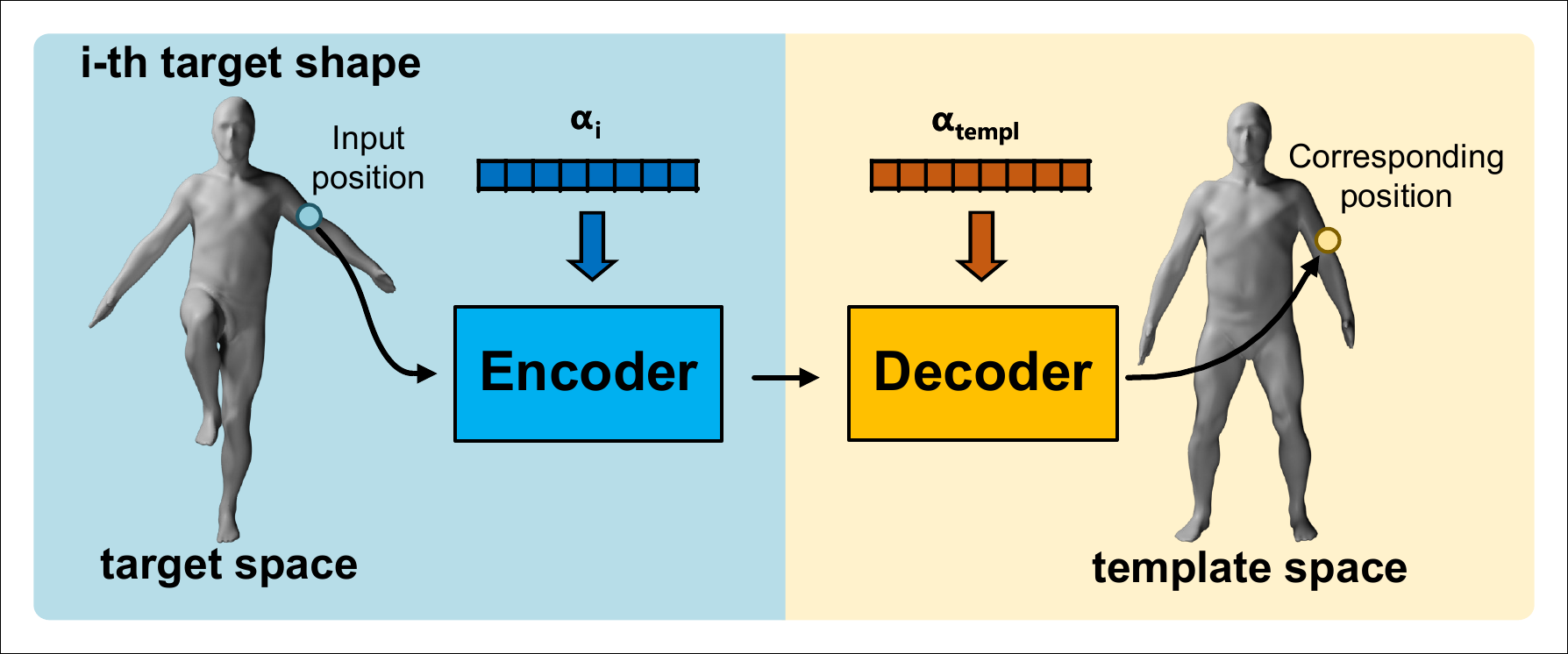}
    \caption{Illustration of our shape representation network. }
    \label{fig:pipeline}
\end{figure}

Our method learns a neural representation for a deformable object exhibited in a collection of shapes represented as signed distance fields (SDF).
Inspired by DIF \cite{Deng_2021_CVPR}, we formulate the deformable object as a template shape encoded in an implicit neural network $\Phi$, and dense correspondence fields $D_{i\rightarrow tmpl}$ predicted by a network ($\mathbb{R}^3 \rightarrow \mathbb{R}^3$) from arbitrary deformation $S_i$ (\ie, target space) to the template space $tmpl$.
Therefore, to reconstruct the target shape $S_i$, signed distance values for arbitrary 3D location $\mathbf{p}$ are queried from the template space via the dense correspondence as
\begin{equation}
    SDF_i(\mathbf{p}) = {\rm \Phi}(D_{i\rightarrow tmpl}(\mathbf{p})).
\end{equation}

With the SDF of a target shape, the 3D mesh can be extracted using surface reconstruction algorithms such as Marching Cubes~\cite{lorensen1987marching}.
Though the overall representation is straightforward, our method focuses on shapes with large deformation, such as moving humans and animals, while DIF~ \cite{Deng_2021_CVPR} can only deal with static categories, such as cars and chairs, we show later in this section effective learning in observations of large deformation is non-trivial.

\subsection{Embedded Shapes and Template}
In this section, we investigate how to learn a reasonable template field.
The previous template field network \cite{Deng_2021_CVPR} did not enforce the template shape to be a valid shape of any subject. 
In practice, this method often generates template with many floating artifacts when dealing with shapes with large deformation (Figure~\ref{fig:template}).
The floating artifacts can mislead the network to find wrong correspondences on template field. However, an ideal template field should have common shape pattern of the target shapes to provide key clues for correspondence. 

In order to learn a reasonable template field, we propose to constrain the template shape in the same latent space with the training shape examples.
To this end, we extend the neural implicit SDF function $\Phi$ to condition on a latent code $\alpha$, where each training shape and the template shape are mapped to an unique latent code.
In this way, the training shape collection naturally forms a strong regularization to ensure a reasonable template shape and effectively prevent flyers.
The latent space also serves naturally for the dense deformation fields between target shape and the template shape.
As illustrated in Figure~\ref{fig:pipeline}, we use the latent codes from the target shape $\alpha_i$ and the template shape $\alpha_{tmpl}$ to drive an encoder and a decoder respectively for dense correspondence prediction. The encoder-decoder network is denoted as $D$.
Inspired the key finding by Simeonov \etal \cite{simeonov2021neural}, that the distance from surface is a key clue to learn 3D correspondence, we add the SDF together with the point location as input of encoder to provide geometry clues.
Therefore, the target shape can be reconstructed as
\begin{equation}
    SDF_i(\mathbf{p}) = {\rm \Phi}\Big(D\big(\mathbf{p},\Phi(\mathbf{p}|\mathbf{\alpha}_i)|\mathbf{\alpha}_i, \mathbf{\alpha}_{tmpl}\big)|\mathbf{\alpha}_{tmpl}\Big).
\label{eq:represent}
\end{equation}
For simplicity, we use $D_{i\rightarrow tmpl}(\mathbf{p})$ for $D(\mathbf{p},{\rm \Phi}(\mathbf{p}|\mathbf{\alpha}_i)|$ $\mathbf{\alpha}_i,\mathbf{\alpha}_{tmpl})$ 
in the following sections.

\subsection{Local Rigid Constraint}
\label{local}
\begin{figure}[th]
    \centering
    \includegraphics[width=0.5\textwidth]{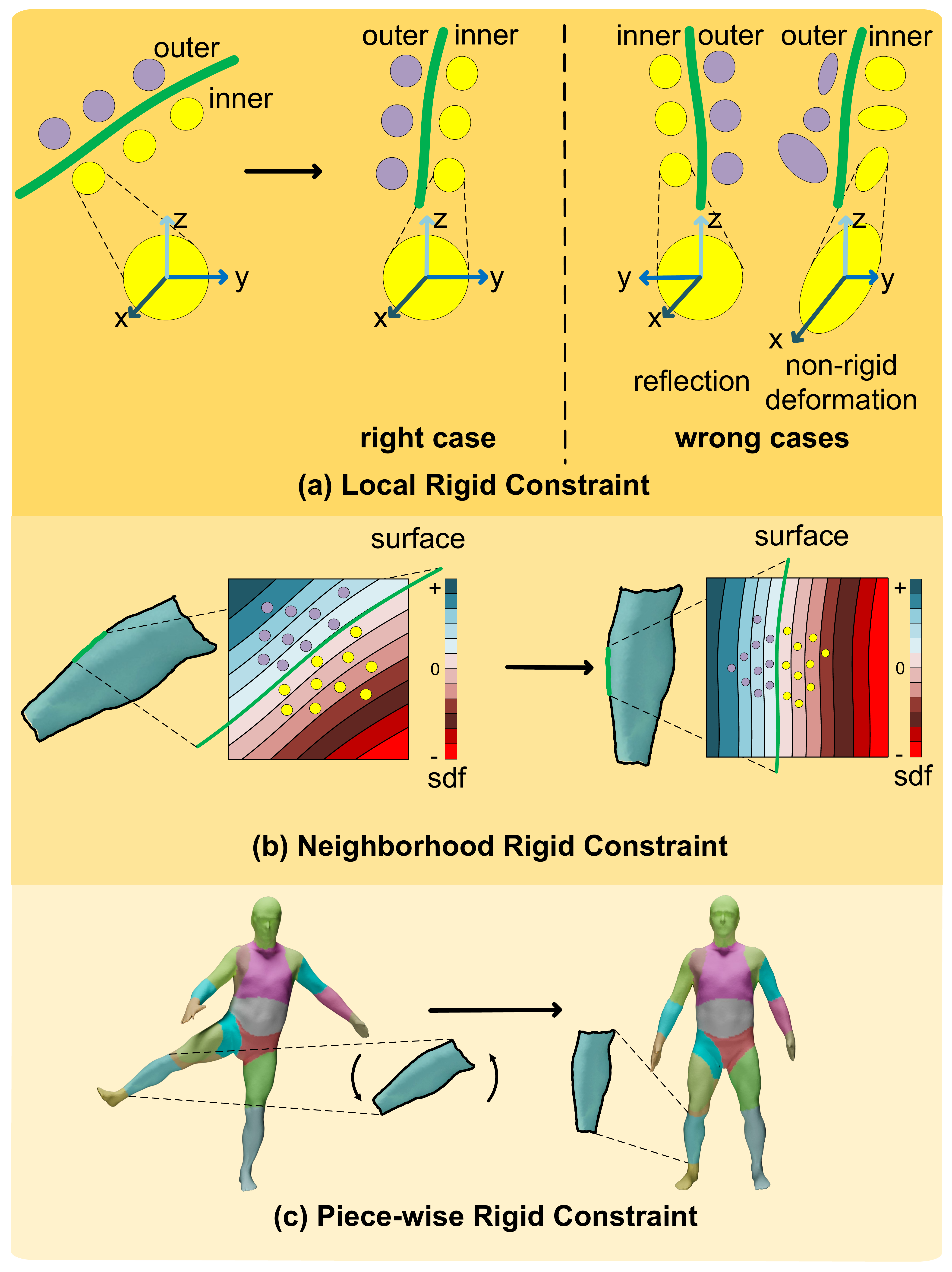}
    \caption{Illustration of hierarchical rigid constraints.}
    \label{fig:rigid}
\end{figure}

During the training, $\Phi$, $D$, and latent code space $\{\alpha\}$ are optimized jointly by minimizing the reconstruction loss on the training shape collection.
However, even with the regularization on the template shape, this is still a highly under-constrained optimization problem, and additional regularization is needed.
A common option is to assume rigid motion in local infinitesimal scopes, and ARAP \cite{alexa2000rigid,sorkine2007rigid} is a widely adopted solution on 3D surfaces.
In this section, we propose, for the first time, a novel ARAP equivalent constraint defined in implicit representation. 

Inspired by Nerfies~\cite{park2021nerfies}, local deformation can be regularized by constraining the singular values from the Jacobian matrix of the deformation field.
In their work, all three singular values are encouraged to be close to 1. 
However, this does not rule out the reflection as part of the rotation \cite{arun1987least}, as illustrated in Figure \ref{fig:rigid} (a), which will deform the shape inside out or erroneously map the symmetry geometry, \eg, left hand in one shape maps to the right hand in another.
While rarely observed in Nerfies~\cite{park2021nerfies} since the deformation in their data is relatively small, this drawback becomes vital when deformations are large.

Our ARAP equivalent constraint is also achieved by constraining the singular values of the Jacobian matrix of the deformation field. The analysis on ARAP equivalence can be found in the supplementary material.
In theory, preference over local rigid deformation is equivalent as encouraging the Jacobian matrix $J(D_{i\rightarrow tmpl})$ to be a rotation matrix \cite{park2021nerfies}.
According to Umeyama \cite{umeyama1991least}, the closest orthogonal matrix of $J(D_{i\rightarrow tmpl})$ in the Froebenius norm is $\mathbf{R}=\mathbf{U}\mathbf{S}\mathbf{V}^T$ with $\mathbf{S}=diag(1,1,det(\mathbf{U}\mathbf{V}^T))$, where $\mathbf{U}, \mathbf{V}$ are obtained via singular value decomposition (SVD), \ie $J(D_{i\rightarrow tmpl})= \mathbf{U}\mathbf{\Sigma}\mathbf{V}^T$ (See more detail in the supplementary material).
Therefore, denoting $\sigma_1$, $\sigma_2$, $\sigma_3$ to be singular values of Jacobian in point $\mathbf{p}$, \ie $J(D_{i\rightarrow tmpl})(\mathbf{p})$, we define the ARAP loss in local region as
\begin{equation}
\begin{aligned}
    L_{arap}=&smoothL1(\sigma_1,1) + smoothL1(\sigma_2,1) \\ &+smoothL1(\sigma_3,det(\mathbf{U}\mathbf{V}^T)).
\end{aligned}
\end{equation}

Note that the reflection happens when $det(\mathbf{U}\mathbf{V}^T)<0$, which is penalized in our ARAP loss since $\sigma_3$ is always positive.
To further penalize reflection, we also directly penalize negative $det(J(D_{i\rightarrow tmpl}))$.
The overall local rigid loss is defined as
\begin{equation}
    L_{lr}=\sum_i\sum_{\mathbf{p}\in S_i^-\cup S_i^0}L_{arap}(\mathbf{p})+relu\Big(-det\big(J(D_{i\rightarrow tmpl})(\mathbf{p})\big)\Big),
\label{eq:lr}
\end{equation}
where $S_i^-$ is the points from shape interior, \ie, $SDF(p)<0$, and $S_i^0$ is the points from shape surface, \ie, $SDF(p)=0$. We obtain the Jacobian matrix via auto gradient mechanism, and use a smooth L1 loss for stable training.

\subsection{Hierarchical Rigid Constraint}
Existing implicit learning methods \cite{park2021nerfies,Deng_2021_CVPR}
only supervise local rigidity of deformation at each point.
However, these methods do not leverage spatial context (\ie, semantic parts, neighborhood distribution) of shapes effectively. 
To mitigate the correspondence ambiguity with spatial context, we propose a new implicit-based hierarchical rigid constraint that consists of three terms at different levels, \ie, the local rigid constraint in infinitesimal scopes (Sec.~\ref{local}), a neighborhood rigid constraint for nearby region, and a piece-wise rigid constraint for large part.

\vspace{1mm}
\noindent \textbf{Neighborhood Rigid Constraint.}
This constraint is applied on small regions but in larger scale than $L_{lr}$, constraining the implicit field in each region to remain consistent after transformation. 
As shown in Figure~\ref{fig:rigid}(b), if points in a small neighborhood undergo the rigid transformation, the SDF of each point near the subject's surface will remain unchanged during deformation.

To this end, we add constraints respecting the above-mentioned property between the template space and each target deformation space.
For each point $\mathbf{p}$ on the surface $S_i^0$ of target deformation, we sample points around it using a Gaussian distribution with $\sigma$ (set to 0.05 in the experiment).
We then estimate the local rotation $\mathbf{R}$ around $\mathbf{p}$ from $J(D_{i\rightarrow tmpl})$ similar to Sec.~\ref{local}.
Each point sampling $\mathbf{p}+\mathbf{\eta}$ ($\mathbf{\eta} \in \mathbb{R}^3$) in the neighborhood is then mapped to the template space at $\mathbf{p}_i^{tmpl} = D_{i\rightarrow tmpl}(\mathbf{p})+\mathbf{R}\mathbf{\eta}$.
We propose a neighborhood rigid loss that penalizes inconsistent SDF values sampled from the target and template space as
\begin{equation}
    L_{nbr}=\sum_i\sum_{\mathbf{p}\in S_i^0} \mathbb{E}_{\mathbf{\eta}\sim\mathcal N(\mathbf{0},\sigma)} \Vert{\rm \Phi}(\mathbf{p}_i^{tmpl}|\alpha_{tmpl})-{\rm \Phi}(\mathbf{p}+\mathbf{\eta}|\alpha_i)\Vert_2^2,
    \label{eq:neighborhood_rigid}
\end{equation}
where $\mathbb{E}$ is the expectation over Gaussian sampling,  
which is implemented by averaging the deviation of SDF values of sampled points. 
Taking numerical stability into account, we follow Levinson \etal \cite{levinson2020analysis} to calculate gradient of $\mathbf{R}$.

\vspace{1mm}
\noindent \textbf{Piece-wise Rigid Constraint.}
In fact, rigid motion can happen not only locally but also in a much larger semantic scope \cite{zuffi2015stitched,li2019lbs}, such as the limb on human body (Figure \ref{fig:rigid} (c)).
These large but rigid structures are often the source of large deformation in 3D space.
We thus add a loss term, named piece-wise rigid loss, to favor rigid motion in large scale to help detect the existence of large rigid parts if any.

\begin{figure}[t]
\centering
\includegraphics[width=0.5\textwidth]{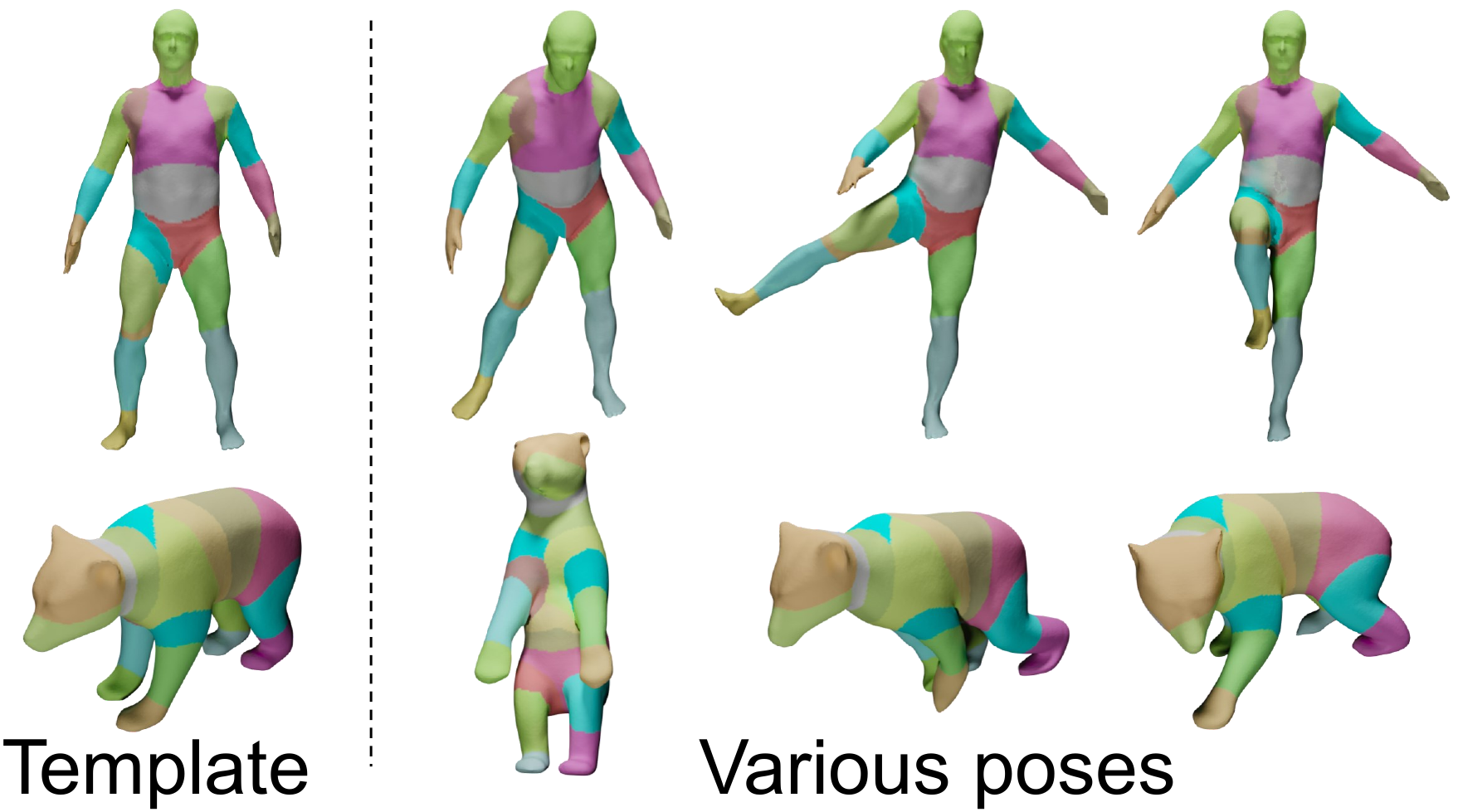}
\caption{Part classification and shapes of two subjects. }
\label{fig:part}
\end{figure}

Our piece-wise rigid loss is enabled via part classification networks that predicts for each 3D point, inside or on the surface of the 3D shape, the probability belonging to each of $N_P$ parts.
With the predicted part association, the least square solution of rigid transformation $(\mathbf{R}_h,\mathbf{t}_h)$ for each part $h$ can be obtained.
The piece-wise rigid loss then penalizes 
the sum of minimal rigid transformation error over points of all parts as
\begin{equation}
\begin{aligned}
&L_{pr}=\sum_i \sum_{h\in[1,N_P]}\min_{\mathbf{R}_h,\mathbf{t}_h}\sum_{\mathbf{p}\in{S_i^0\cup S_i^-}}
    \mathbf{P}_{h}(\mathbf{p}) \Vert (\mathbf{R}_h\mathbf{p}+\mathbf{t}_h)\\
&    -{D_{i \rightarrow tmpl}(\mathbf{p})}\Vert_2^2,
\end{aligned}    
\label{eq:rigidloss}
\end{equation}
where 
$\mathbf{P}_h$ is the predicted probability belonging to part $h$. 
In general, the points from shape interior have strong correlation with surface points. Therefore, we not only constrain surface points, but also inner points. 
In practice, the part classification network is learned in self-supervised way, jointly with $D_{i \rightarrow tmpl}$, which needs only a pre-defined total number of parts $N_P$. 
Because the calculation of per-part rigid transformation $(\mathbf{R}_h,\mathbf{t}_h)$ could be slow, 
we therefore leverage a closed-form analytical solution to get the minimal rigid transformation error 
following Sorkine-Hornung \etal \cite{sorkine2017least} for efficient and differentiable implementation of Eq.~\ref{eq:rigidloss}.
Figure \ref{fig:part} shows that our method can effectively learn part classification in self-supervised manner.

Overall, our hierarchical rigid constraint is defined as
\begin{equation}
    L_{rigid} = w_{lr}L_{lr}+w_{nbr}L_{nbr}+w_{pr}L_{pr}
\end{equation}
where $w_{lr}$, $w_{nbr}$ and $w_{pr}$ are loss weights. 

\subsection{Implementation Details}
We train our model in an end-to-end manner, and the latent code and parameters of all networks are optimized together during training. 

Besides the hierarchical rigid constraint, we also add a loss term to incorporate directly supervision from the training shapes on the SDF as
\begin{equation}
\begin{aligned}
     &L_{sdf}=\\
     &\sum_i\Big(w_s\sum_{\mathbf{p}\in S_i}\vert {\rm \Phi}(\mathbf{p}|\alpha_i) - \bar{s}\vert + w_n\sum_{\mathbf{p}\in S_i^0}(1-S_c(\nabla\Phi(\mathbf{p}|\alpha_i),\bar{\mathbf{n}})\Big)\\
             &+w_{Eik}\sum_{\mathbf{p}\in S_i}\vert \Vert \nabla{\rm \Phi}(\mathbf{p}|\alpha_i) \Vert_2-1 \vert+w_{\rho}\sum_{\mathbf{p}\in S_i\backslash S_i^0}\rho({\rm \Phi}(\mathbf{p})|\alpha_i)),
\end{aligned}
\end{equation}
where $S_c$ is cosine similarity, $\bar{{s}}$ is ground truth SDF value, and $\bar{\mathbf{n}}$ is ground truth normal.
The first term directly supervises the SDF value, the second term supervises the surface normal, and the third term regularizes the amplitude of SDF gradient to satisfy Eikonal equation. The fourth term refrains from the off-surface points with SDF values close to 0, where $\rho(s) = exp(-\delta \cdot \vert s \vert), s\gg1$.
We use the similar method to supervise queried SDF values (Eq.~\ref{eq:represent}), more details can be found in supplementary material.

We also supervise the surface normal consistency jointly with the deformation field.
Specifically, we rotate the ground truth  $\bar{\mathbf{n}}$ from the target space to the template space using predicted rotation $J(D_{i\rightarrow tmpl})(\mathbf{p})$, i.e., $J(D_{i\rightarrow tmpl})(\mathbf{p})\bar{\mathbf{n}}$\cite{lee2013smooth}, and then compare it with the normal directly estimated with SDFs of correspondence in the template space, i.e., $\nabla_{D_{i\rightarrow tmpl}}{\rm \Phi}(D_{i\rightarrow tmpl}(\mathbf{p})|\alpha_{tmpl})$. 
The normal loss $L_{pfn}$ is defined as 
\begin{equation}
\begin{aligned}
    L_{pfn}=\sum_i\sum_{\mathbf{p}\in S_i^0}(1-&S_c(\nabla_{D_{i\rightarrow tmpl}}{\rm \Phi}(D_{i\rightarrow tmpl}(\mathbf{p})|\alpha_{tmpl}),\\
    &J(D_{i\rightarrow tmpl})(\mathbf{p})\bar{\mathbf{n}} ))
\end{aligned}
\end{equation}

We also use regularization terms on latent codes $L_{reg}=\sum_i\Vert \mathbf{\alpha}_i\Vert$ and enforce the latent code of template to be close to its nearest latent code of shape in the training set.

Inspired by 3D-CODED~\cite{groueix2018b} that pre-trains the network to enforce the predicted correspondence of a point in input shape to close to the input point, we use a loss term  $L_{recon}$ to enforce the self-correspondence $D_{i\rightarrow i}(\mathbf{p})$ of the target shape to close to $\mathbf{p}$ using $L_2$ loss, which enables good initial correspondence. 
The hyper-parameters such as loss weights are fixed over all the experiments.
\begin{table*}[ht]
\resizebox{\linewidth}{!}{
\begin{tabular}{|c|ccc|cccccccccccccccccc|}
\hline
\multirow{3}{*}{} & \multicolumn{3}{c|}{\multirow{2}{*}{D-FAUST\cite{dfaust:CVPR:2017}}}                                                         & \multicolumn{18}{c|}{DeformingThings4D\cite{li20214dcomplete}}                                                                                                                                                                                                                                                                                                                                                                                                                                                                                                                                                                                                                                                                                                                \\ \cline{5-22} 
                  & \multicolumn{3}{c|}{}               & \multicolumn{3}{c|}{bear}                                                                                                  & \multicolumn{3}{c|}{rabbit}                                                                                                & \multicolumn{3}{c|}{elephant}                                                                                              & \multicolumn{3}{c|}{whale}                                                                                                 & \multicolumn{3}{c|}{deer}                                                                                                  & \multicolumn{3}{c|}{average}                                                                          \\ \cline{2-22} 
                  & \multicolumn{1}{c|}{CD$\downarrow$} & \multicolumn{1}{c|}{IoU$\uparrow$}  & \textit{corr$\downarrow$} & \multicolumn{1}{c|}{CD$\downarrow$} & \multicolumn{1}{c|}{IoU$\uparrow$}  & \multicolumn{1}{c|}{\textit{corr$\downarrow$}} & \multicolumn{1}{c|}{CD$\downarrow$} & \multicolumn{1}{c|}{IoU$\uparrow$}  & \multicolumn{1}{c|}{\textit{corr$\downarrow$}} & \multicolumn{1}{c|}{CD$\downarrow$} & \multicolumn{1}{c|}{IoU$\uparrow$}  & \multicolumn{1}{c|}{\textit{corr$\downarrow$}} & \multicolumn{1}{c|}{CD$\downarrow$} & \multicolumn{1}{c|}{IoU$\uparrow$}  & \multicolumn{1}{c|}{\textit{corr$\downarrow$}} & \multicolumn{1}{c|}{CD$\downarrow$} & \multicolumn{1}{c|}{IoU$\uparrow$}  & \multicolumn{1}{c|}{\textit{corr$\downarrow$}} & \multicolumn{1}{c|}{CD$\downarrow$} & \multicolumn{1}{c|}{IoU$\uparrow$}  & \textit{corr$\downarrow$} \\ \hline
DIF               & \multicolumn{1}{c|}{11.936}          & \multicolumn{1}{c|}{0.647}          & 0.0901                    & \multicolumn{1}{c|}{16.579}         & \multicolumn{1}{c|}{0.636}          & \multicolumn{1}{c|}{0.2062}                    & \multicolumn{1}{c|}{14.005}         & \multicolumn{1}{c|}{0.566}          & \multicolumn{1}{c|}{0.1515}                    & \multicolumn{1}{c|}{187.662}        & \multicolumn{1}{c|}{0.431}          & \multicolumn{1}{c|}{0.0612}                    & \multicolumn{1}{c|}{22.146}         & \multicolumn{1}{c|}{0.492}          & \multicolumn{1}{c|}{0.0812}                    & \multicolumn{1}{c|}{24.772}         & \multicolumn{1}{c|}{0.489}          & \multicolumn{1}{c|}{0.1948}                    & \multicolumn{1}{c|}{44.053}         & \multicolumn{1}{c|}{0.547}          & 0.1303                    \\ \hline
3D-CODED          & \multicolumn{1}{c|}{3.450}          & \multicolumn{1}{c|}{0.592}          & 0.1038                    & \multicolumn{1}{c|}{1.327}          & \multicolumn{1}{c|}{0.824}          & \multicolumn{1}{c|}{0.1654}                    & \multicolumn{1}{c|}{\textbf{1.303}} & \multicolumn{1}{c|}{0.797}          & \multicolumn{1}{c|}{0.1271}                    & \multicolumn{1}{c|}{0.832}          & \multicolumn{1}{c|}{0.888}          & \multicolumn{1}{c|}{0.0377}                    & \multicolumn{1}{c|}{2.689}          & \multicolumn{1}{c|}{0.692}          & \multicolumn{1}{c|}{0.0710}                    & \multicolumn{1}{c|}{2.683} & \multicolumn{1}{c|}{0.656}          & \multicolumn{1}{c|}{0.1440}                    & \multicolumn{1}{c|}{1.658}          & \multicolumn{1}{c|}{0.784}          & 0.0688                    \\ \hline
Our               & \multicolumn{1}{c|}{\textbf{1.594}} & \multicolumn{1}{c|}{\textbf{0.881}} & \textbf{0.0304}           & \multicolumn{1}{c|}{\textbf{0.439}} & \multicolumn{1}{c|}{\textbf{0.940}} & \multicolumn{1}{c|}{\textbf{0.0700}}           & \multicolumn{1}{c|}{1.731}          & \multicolumn{1}{c|}{\textbf{0.897}} & \multicolumn{1}{c|}{\textbf{0.0925}}           & \multicolumn{1}{c|}{\textbf{0.587}} & \multicolumn{1}{c|}{\textbf{0.910}} & \multicolumn{1}{c|}{\textbf{0.0198}}           & \multicolumn{1}{c|}{\textbf{1.754}} & \multicolumn{1}{c|}{\textbf{0.908}} & \multicolumn{1}{c|}{\textbf{0.0175}}           & \multicolumn{1}{c|}{\textbf{2.121}}          & \multicolumn{1}{c|}{\textbf{0.870}} & \multicolumn{1}{c|}{\textbf{0.0786}}           & \multicolumn{1}{c|}{\textbf{1.165}} & \multicolumn{1}{c|}{\textbf{0.912}} & \textbf{0.0557}           \\ \hline
\end{tabular}
}
\caption{Capacity evaluation on D-FAUST and DeformingThings4D.}
\label{tab:capacity}
\end{table*}

\begin{table}[ht]
\centering
\resizebox{0.5\linewidth}{!}{
\begin{tabular}{|c|ccc|}
\hline
\multirow{2}{*}{}              & \multicolumn{3}{c|}{\multirow{1}{*}{MANO\cite{romero2022embodied}}}                                                            \\\cline{2-4} 
                               & \multicolumn{1}{c|}{CD$\downarrow$} & \multicolumn{1}{c|}{IoU$\uparrow$}  & \textit{corr$\downarrow$} \\ \hline
DIF                            & \multicolumn{1}{c|}{8.137}          & \multicolumn{1}{c|}{0.824}          & 0.0557                    \\ \hline
\multicolumn{1}{|l|}{3D-CODED} & \multicolumn{1}{c|}{0.833}          & \multicolumn{1}{c|}{0.879}          & 0.0241                    \\ \hline
Our                            & \multicolumn{1}{c|}{\textbf{0.150}} & \multicolumn{1}{c|}{\textbf{0.935}} & \textbf{0.0024}           \\ \hline
\end{tabular}
}
\caption{Capacity evaluation on generated MANO\cite{romero2022embodied} hands.}
\label{tab:mano}
\end{table}



\section{Experiment}
    \begin{figure}[ht]
    \centering
     \includegraphics[width=0.5\textwidth]{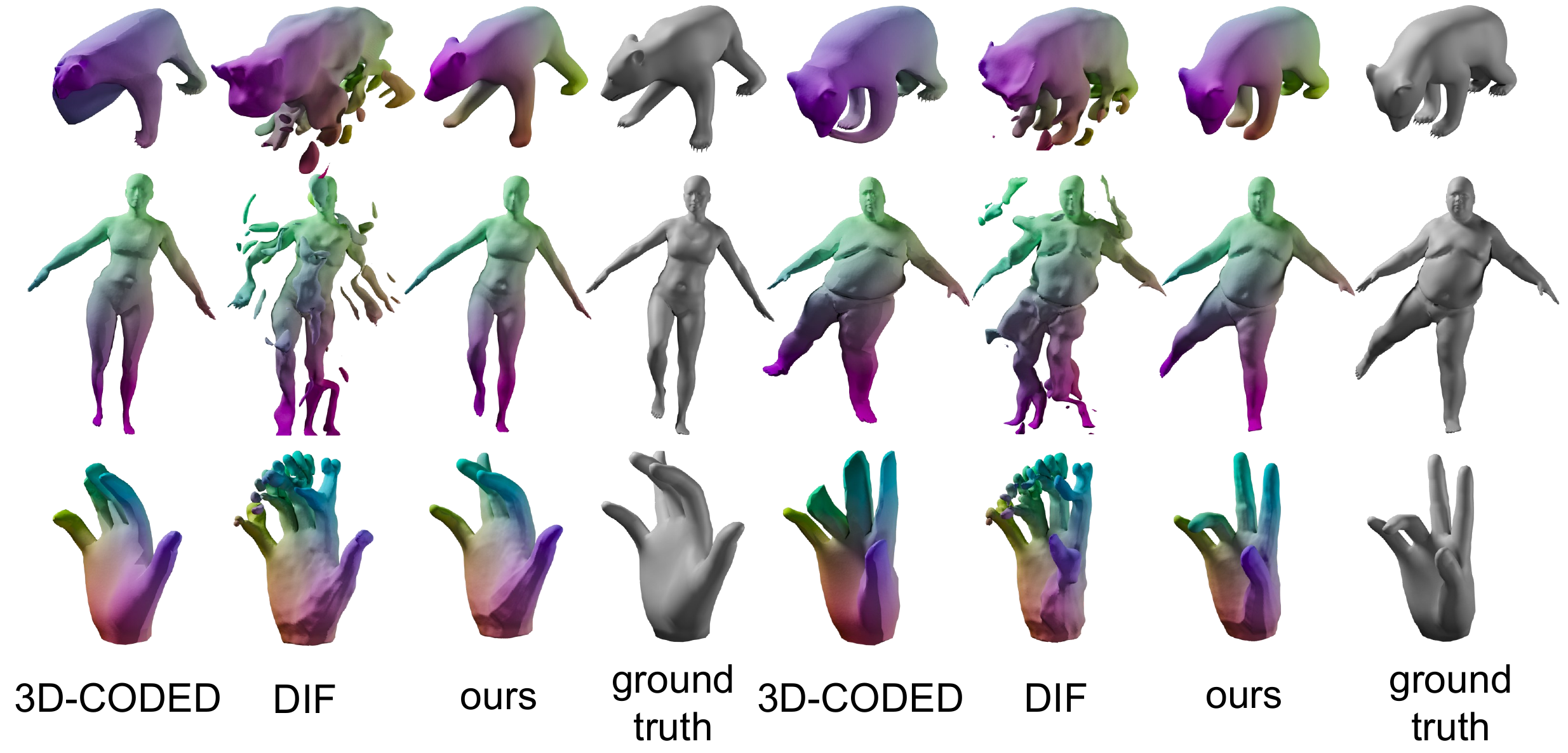}
     \caption{Comparison of model representation capability from our method with DIF and 3D-CODED. 
     Our method outperforms DIF and 3D-CODED by a large margin on the representation capability. 
    }
    \label{fig:main_train_compare}
\end{figure}

\begin{figure}[ht]
    \centering
     \includegraphics[width=0.5\textwidth]{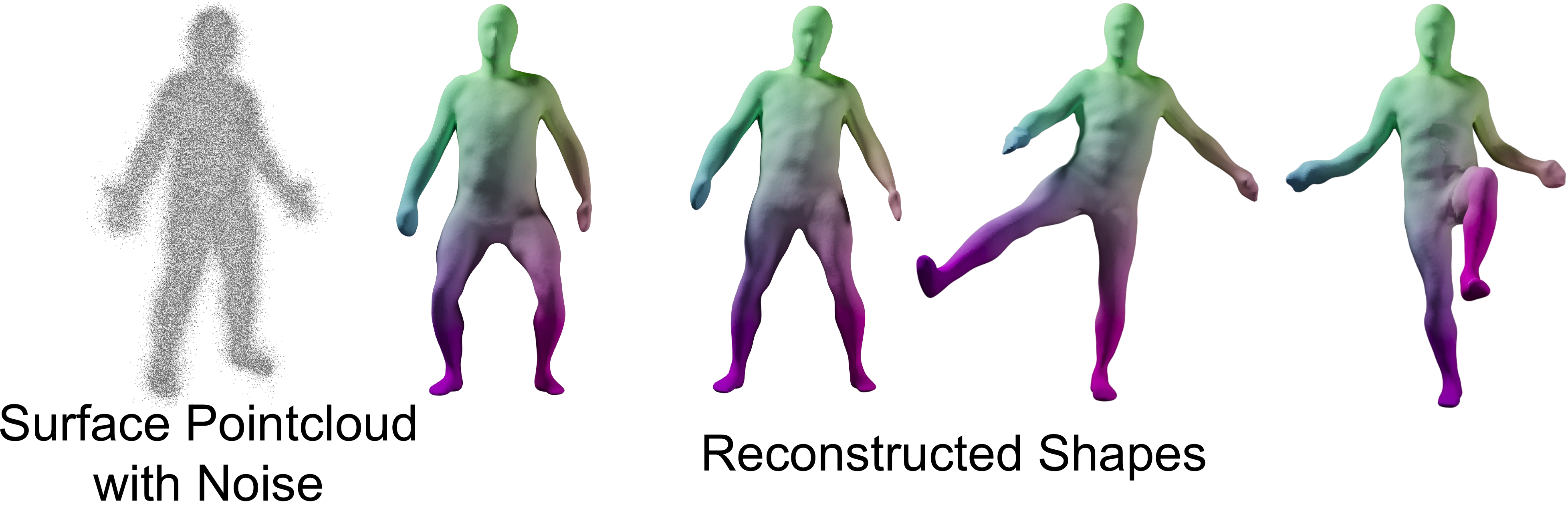}
     \caption{Results on datasets with synthesized noise. 
     We can observe that our method can generate reasonable results with synthesized noise.
    }
    \label{fig:noise}
\end{figure}

\subsection{Dataset}
We use the human dataset D-FAUST~\cite{dfaust:CVPR:2017} and animal dataset DeformingThings4D~\cite{li20214dcomplete} for evaluation. 
D-FAUST contains 5 males and 5 females. Each person performs various movements, such as punching and waving arms. We use the same data split as Atzmon \etal~\cite{atzmon2020sal}. 
Shapes from several sequences are randomly split.
Although pose is available in the dataset, it is not used in our experiments.
DeformingThings4D contains various animals. In our experiment, we select 5 animals with very different shapes and skeleton structures, including bear, bunny, whale, elephant and deer.

\subsection{Representation Capability}
We evaluate the representation capability of our method by comparing reconstructed shapes in the training set with SOTA shape representation methods, DIF \cite{Deng_2021_CVPR} and 3D-CODED \cite{groueix2018b}.
To get the results with DIF, we only use the queried SDF value from template field via correspondence without per-point correction to reconstruct shapes, because shapes with correction term are not just a deformed template, their geometry mainly depends on correction instead of correspondence.
Table~\ref{tab:capacity} shows quantitative comparison on D-FAUST and DeformingThings4D, measured with Chamfer distance (CD$\times1000$) and Intersection over Union (IoU).
On D-FAUST, our method outperforms 3D-CODED and DIF.
On DeformingThing4D, our method also outperforms DIF and achieves better IoU than 3D-CODED with comparable average performance on CD. 
We find that in our task IoU is a more stable evaluation metric than CD, because CD is sensitive to small floating components.
In the supplementary material, we show several failure cases where floating components make CD increase.
We also test our method on synthesized MANO \cite{romero2022embodied} dataset. As shown in Table~\ref{tab:mano} and Figure~\ref{fig:main_train_compare}, our model can deal with challenging hand shapes and outperforms 3D-CODED and DIF.
We futher demonstrate the robustness of our method against noise. We follow DeepSDF \cite{park2019deepsdf} to apply synthesized noise on the depth maps where $\sigma=0.01$, before calculating SDFs from these depths. Although suffering from heavy noises, our model still performs well. Figure~\ref{fig:noise} demonstrates the qualitative experiment, and for the quantitative experiment: Chamfer=0.936, IoU=0.830.

\begin{table*}[ht]
\centering
\resizebox{\linewidth}{!}{
\begin{tabular}{|c|ccc|cccccccccccccccccc|}
\hline
\multirow{3}{*}{} & \multicolumn{3}{c|}{\multirow{2}{*}{D-FAUST\cite{dfaust:CVPR:2017}}}                                   & \multicolumn{18}{c|}{DeformingThings4D\cite{li20214dcomplete}}                                                                                                                                                                                                                                                                                                                                                                                                                                                                                                                                                                                                                                                                                          \\ \cline{5-22} 
                  & \multicolumn{3}{c|}{}               & \multicolumn{3}{c|}{bear}                                                                                                  & \multicolumn{3}{c|}{rabbit}                                                                                                & \multicolumn{3}{c|}{elephant}                                                                                              & \multicolumn{3}{c|}{whale}                                                                                                 & \multicolumn{3}{c|}{deer}                                                                                                  & \multicolumn{3}{c|}{average}                                                                          \\ \cline{2-22} 
                  & \multicolumn{1}{c|}{CD$\downarrow$} & \multicolumn{1}{c|}{IoU$\uparrow$}  & \textit{corr$\downarrow$} & \multicolumn{1}{c|}{CD$\downarrow$} & \multicolumn{1}{c|}{IoU$\uparrow$}  & \multicolumn{1}{c|}{\textit{corr$\downarrow$}} & \multicolumn{1}{c|}{CD$\downarrow$} & \multicolumn{1}{c|}{IoU$\uparrow$}  & \multicolumn{1}{c|}{\textit{corr$\downarrow$}} & \multicolumn{1}{c|}{CD$\downarrow$} & \multicolumn{1}{c|}{IoU$\uparrow$}  & \multicolumn{1}{c|}{\textit{corr$\downarrow$}} & \multicolumn{1}{c|}{CD$\downarrow$} & \multicolumn{1}{c|}{IoU$\uparrow$}  & \multicolumn{1}{c|}{\textit{corr$\downarrow$}} & \multicolumn{1}{c|}{CD$\downarrow$} & \multicolumn{1}{c|}{IoU$\uparrow$}  & \multicolumn{1}{c|}{\textit{corr$\downarrow$}} & \multicolumn{1}{c|}{CD$\downarrow$} & \multicolumn{1}{c|}{IoU$\uparrow$}  & \textit{corr$\downarrow$} \\ \hline
DIF               & \multicolumn{1}{c|}{11.790}         & \multicolumn{1}{c|}{0.636}          & 0.0917                    & \multicolumn{1}{c|}{17.010}         & \multicolumn{1}{c|}{0.629}          & \multicolumn{1}{c|}{0.1959}                    & \multicolumn{1}{c|}{15.057}         & \multicolumn{1}{c|}{0.548}          & \multicolumn{1}{c|}{0.1648}                    & \multicolumn{1}{c|}{192.317}        & \multicolumn{1}{c|}{0.425}          & \multicolumn{1}{c|}{0.0617}                    & \multicolumn{1}{c|}{22.247}         & \multicolumn{1}{c|}{0.486}          & \multicolumn{1}{c|}{0.0851}                    & \multicolumn{1}{c|}{28.907}         & \multicolumn{1}{c|}{0.426}          & \multicolumn{1}{c|}{0.2360}                    & \multicolumn{1}{c|}{45.672}         & \multicolumn{1}{c|}{0.531}          & 0.1487                    \\ \hline
3D-CODED          & \multicolumn{1}{c|}{3.389}          & \multicolumn{1}{c|}{0.597}          & 0.1068                    & \multicolumn{1}{c|}{1.276}          & \multicolumn{1}{c|}{0.826}          & \multicolumn{1}{c|}{0.1720}                    & \multicolumn{1}{c|}{\textbf{1.467}} & \multicolumn{1}{c|}{0.791}          & \multicolumn{1}{c|}{0.1326}                    & \multicolumn{1}{c|}{0.807}          & \multicolumn{1}{c|}{0.889}          & \multicolumn{1}{c|}{0.0360}                    & \multicolumn{1}{c|}{2.580}          & \multicolumn{1}{c|}{0.697}          & \multicolumn{1}{c|}{0.0799}                    & \multicolumn{1}{c|}{\textbf{2.823}} & \multicolumn{1}{c|}{0.657}          & \multicolumn{1}{c|}{0.1631}                    & \multicolumn{1}{c|}{1.669}          & \multicolumn{1}{c|}{0.784}          & 0.0742                    \\ \hline
Our               & \multicolumn{1}{c|}{\textbf{1.480}} & \multicolumn{1}{c|}{\textbf{0.890}} & \textbf{0.0307}           & \multicolumn{1}{c|}{\textbf{0.609}} & \multicolumn{1}{c|}{\textbf{0.940}} & \multicolumn{1}{c|}{\textbf{0.0789}}           & \multicolumn{1}{c|}{1.716}          & \multicolumn{1}{c|}{\textbf{0.891}} & \multicolumn{1}{c|}{\textbf{0.0831}}           & \multicolumn{1}{c|}{\textbf{0.720}} & \multicolumn{1}{c|}{\textbf{0.909}} & \multicolumn{1}{c|}{\textbf{0.0216}}           & \multicolumn{1}{c|}{\textbf{1.613}} & \multicolumn{1}{c|}{\textbf{0.914}} & \multicolumn{1}{c|}{\textbf{0.0241}}           & \multicolumn{1}{c|}{2.917}          & \multicolumn{1}{c|}{\textbf{0.867}} & \multicolumn{1}{c|}{\textbf{0.1003}}           & \multicolumn{1}{c|}{\textbf{1.321}} & \multicolumn{1}{c|}{\textbf{0.911}} & \textbf{0.0616}           \\ \hline
\end{tabular}
}
\caption{Shape reconstruction from full observation.}
\label{tab:full_eval}
\end{table*}

\begin{table*}[ht]
\resizebox{\linewidth}{!}{
\begin{tabular}{|c|ccc|cccccccccccccccccc|}
\hline
\multirow{3}{*}{} & \multicolumn{3}{c|}{\multirow{2}{*}{D-FAUST\cite{dfaust:CVPR:2017}}}                                   & \multicolumn{18}{c|}{DeformingThings4D\cite{li20214dcomplete}}                                                                                                                                                                                                                                                                                                                                                                                                                                                                                                                                                                                                                                                                                          \\ \cline{5-22} 
                  & \multicolumn{3}{c|}{}               & \multicolumn{3}{c|}{bear}                                                                                                  & \multicolumn{3}{c|}{rabbit}                                                                                                & \multicolumn{3}{c|}{elephant}                                                                                              & \multicolumn{3}{c|}{whale}                                                                                                 & \multicolumn{3}{c|}{deer}                                                                                                  & \multicolumn{3}{c|}{average}                                                                          \\ \cline{2-22} 
                  & \multicolumn{1}{c|}{CD$\downarrow$} & \multicolumn{1}{c|}{IoU$\uparrow$}  & \textit{corr$\downarrow$} & \multicolumn{1}{c|}{CD$\downarrow$} & \multicolumn{1}{c|}{IoU$\uparrow$}  & \multicolumn{1}{c|}{\textit{corr$\downarrow$}} & \multicolumn{1}{c|}{CD$\downarrow$} & \multicolumn{1}{c|}{IoU$\uparrow$}  & \multicolumn{1}{c|}{\textit{corr$\downarrow$}} & \multicolumn{1}{c|}{CD$\downarrow$} & \multicolumn{1}{c|}{IoU$\uparrow$}  & \multicolumn{1}{c|}{\textit{corr$\downarrow$}} & \multicolumn{1}{c|}{CD$\downarrow$} & \multicolumn{1}{c|}{IoU$\uparrow$}  & \multicolumn{1}{c|}{\textit{corr$\downarrow$}} & \multicolumn{1}{c|}{CD$\downarrow$} & \multicolumn{1}{c|}{IoU$\uparrow$}  & \multicolumn{1}{c|}{\textit{corr$\downarrow$}} & \multicolumn{1}{c|}{CD$\downarrow$} & \multicolumn{1}{c|}{IoU$\uparrow$}  & \textit{corr$\downarrow$} \\ \hline
DIF               & \multicolumn{1}{c|}{11.787}         & \multicolumn{1}{c|}{0.632}          & 0.0924                    & \multicolumn{1}{c|}{17.177}         & \multicolumn{1}{c|}{0.621}          & \multicolumn{1}{c|}{0.2228}                    & \multicolumn{1}{c|}{15.238}         & \multicolumn{1}{c|}{0.543}          & \multicolumn{1}{c|}{0.1963}                    & \multicolumn{1}{c|}{190.240}        & \multicolumn{1}{c|}{0.428}          & \multicolumn{1}{c|}{0.0878}                    & \multicolumn{1}{c|}{22.255}         & \multicolumn{1}{c|}{0.495}          & \multicolumn{1}{c|}{0.1001}                    & \multicolumn{1}{c|}{27.268}         & \multicolumn{1}{c|}{0.467}          & \multicolumn{1}{c|}{0.2293}                    & \multicolumn{1}{c|}{45.239}         & \multicolumn{1}{c|}{0.534}          & 0.1673                    \\ \hline
Our               & \multicolumn{1}{c|}{\textbf{1.689}} & \multicolumn{1}{c|}{\textbf{0.881}} & \textbf{0.0324}           & \multicolumn{1}{c|}{\textbf{1.757}} & \multicolumn{1}{c|}{\textbf{0.910}} & \multicolumn{1}{c|}{\textbf{0.0922}}           & \multicolumn{1}{c|}{\textbf{2.407}} & \multicolumn{1}{c|}{\textbf{0.868}} & \multicolumn{1}{c|}{\textbf{0.0928}}           & \multicolumn{1}{c|}{\textbf{2.196}} & \multicolumn{1}{c|}{\textbf{0.873}} & \multicolumn{1}{c|}{\textbf{0.0359}}           & \multicolumn{1}{c|}{\textbf{1.685}} & \multicolumn{1}{c|}{\textbf{0.885}} & \multicolumn{1}{c|}{\textbf{0.0237}}           & \multicolumn{1}{c|}{\textbf{3.403}} & \multicolumn{1}{c|}{\textbf{0.833}} & \multicolumn{1}{c|}{\textbf{0.0960}}           & \multicolumn{1}{c|}{\textbf{2.156}} & \multicolumn{1}{c|}{\textbf{0.882}} & \textbf{0.0681}           \\ \hline
\end{tabular}
}
\caption{Shape reconstruction from partial point clouds.}
\label{tab:partial_eval}
\end{table*}

\subsection{Shape Interpolation}
In this section, we demonstrate that our model can also represent shapes similar to the shapes in training set.
We evaluate shape representation capability through model fitting from full observation or partial point cloud rendered from D-FAUST and DeformingThings4D.

In experiments, we obtain each partial point cloud from single depth image, while obtaining each full observation from 20 depth images captured from multiple views.
During evaluation, we use the shapes from training sequences but not involved in training.
With trained models, we conduct model fitting from partial point clouds or full observations by optimizing the latent code $\alpha$ and a global transformation with the following function
\begin{equation}
    L=w_{s}L_{sdf}+w_{s}L_{pbs}+w_n L_{pbn}+w_n L_{pfn}+w_{reg}L_{reg},
\label{eq:eval}
\end{equation}
where $L_{pbs}$ and $L_{pbn}$ are constraints to supervise queried SDF values from the template and normal of queried SDF values. Details can be found in supplementary material. 

\begin{figure}[ht]
    \centering
     \includegraphics[width=0.5\textwidth]{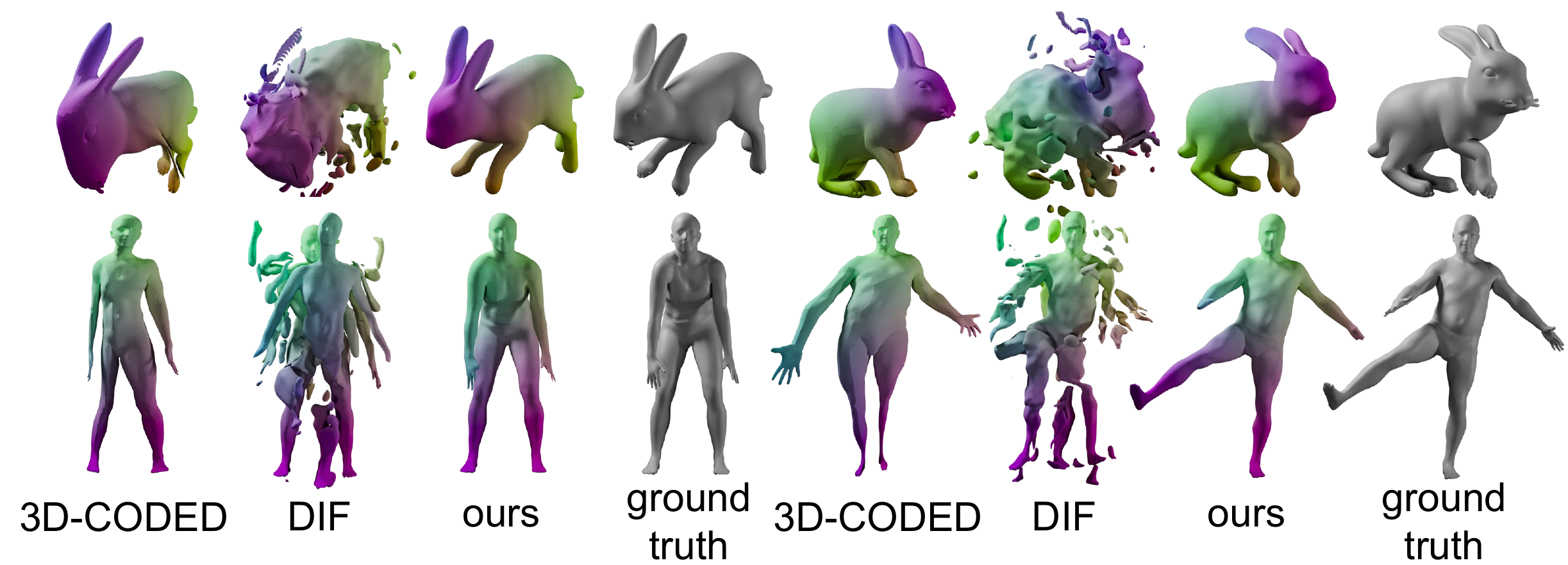}
     \caption{Comparison of shape reconstruction from full point clouds with DIF and 3D-CODED. Our method outperforms DIF and 3D-CODED by a large margin on the reconstruction results. 
    }
    \label{fig:main_full_compare}
\end{figure}

\begin{figure}[ht]
    \centering
     \includegraphics[width=0.5\textwidth]{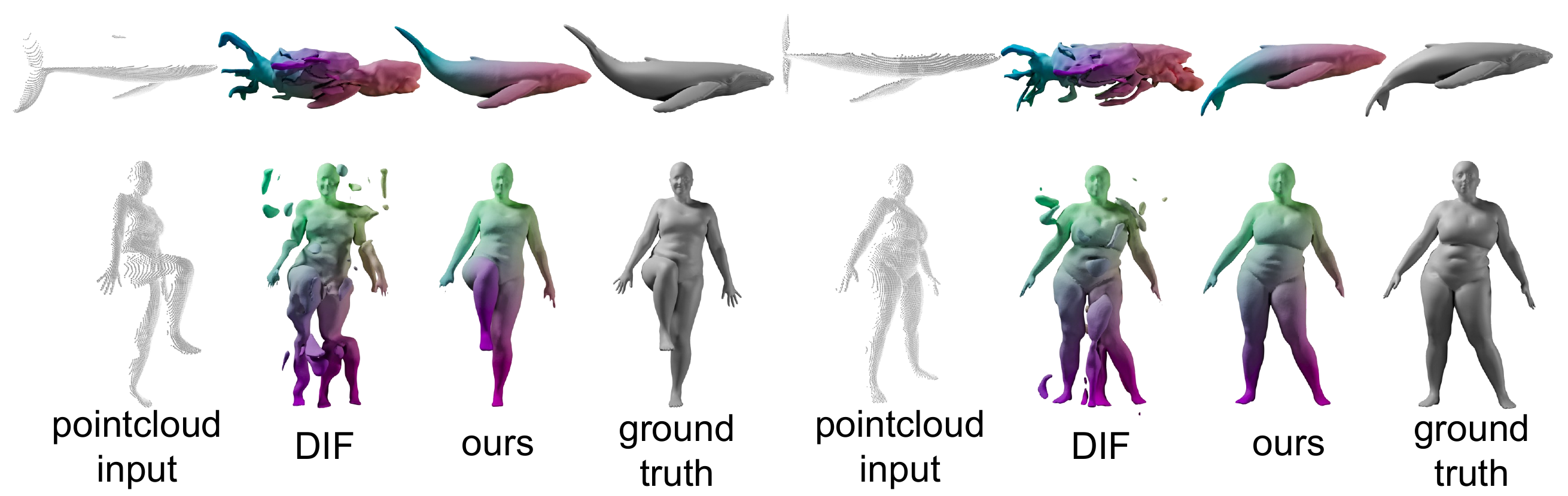}
     \caption{Comparison of shape reconstruction from partial point clouds with DIF and our method. Our method outperforms DIF by a large margin on the reconstruction results. 
    }
    \label{fig:main_partial_compare}
\end{figure}

For full observation input, our method outperforms 3D-CODED on 10 subjects of D-FAUST (Table~\ref{tab:full_eval}). 
As shown in Figure~\ref{fig:main_full_compare}, the reconstructed shapes with 3D-CODED may distort to reduce CD. 
For shape reconstruction from partial observation, we compare our method with the SoTA method DIF~\cite{Deng_2021_CVPR}.
As shown in Figure~\ref{fig:main_partial_compare} and Table~\ref{tab:partial_eval},
our method outperforms DIF by a large margin for partial point clouds. 
Since 3D-CODED fails to work well on partial observation, we do not show its results in Table~\ref{tab:partial_eval}. 

\subsection{Correspondence}
We evaluate the accuracy of correspondence for training shapes (Table~\ref{tab:capacity}), which shows the capacity of our method, and for shapes reconstructed from partial observation (Table~\ref{tab:partial_eval}) and full observation (Table~\ref{tab:full_eval}) of unseen data. 
We compare our method with unsupervised correspondence learning methods: 3D-CODED~\cite{groueix2018b} and DIF~\cite{Deng_2021_CVPR}.
Our method achieves better correspondence performance, \ie, \emph{corr}, than DIF and 3D-CODED.
The correspondence metric \emph{corr} is calculated in the following manner. 
Given the reconstructed shapes $S_{f_1}$ and $S_{f_2}$, we optimize Eq.~\ref{eq:eval} to get deformation field, and then warp $S_{f_1}$ and $S_{f_2}$ to template space. For each point $\mathbf{p}_{f_1}$ on $S_{f_1}$ warped to template, we find the nearest point warped from a point (\eg, $\mathbf{p}_{f_2}$) on $S_{f_2}$ to template, then we set $\mathbf{p}_{f_2}$ on $S_{f_2}$ as the correspondence of $\mathbf{p}_{f_1}$ on $S_{f_1}$. We calculate the geodesic distance between ground truth corresponding point $\bar{\mathbf{p}}_{f_2}$ and predicted corresponding point $\mathbf{p}_{f_2}$ as error. For each subject, we randomly select 100 shapes from the training set and the testing set, and evaluate the geodesic distance error between each pair.
The correspondence metric \emph{corr} is calculated by averaging the geodesic distance error of each pair of shapes.
We also test model trained on noise depths (Figure.~\ref{fig:noise}) and \emph{corr}=0.0233.

\subsection{Template Visualization}
Figure~\ref{fig:template} visualizes the template fields generated by our method and DIF. The template built by our method is in the manifold of human shape, while the template by DIF does not follow human shape, which shows that the same points in differently posed shapes may have different correspondences in template, such as points on arms. Therefore, our method is effective to get reasonable template.
\begin{figure}[ht]
\centering
\includegraphics[width=0.3\textwidth]{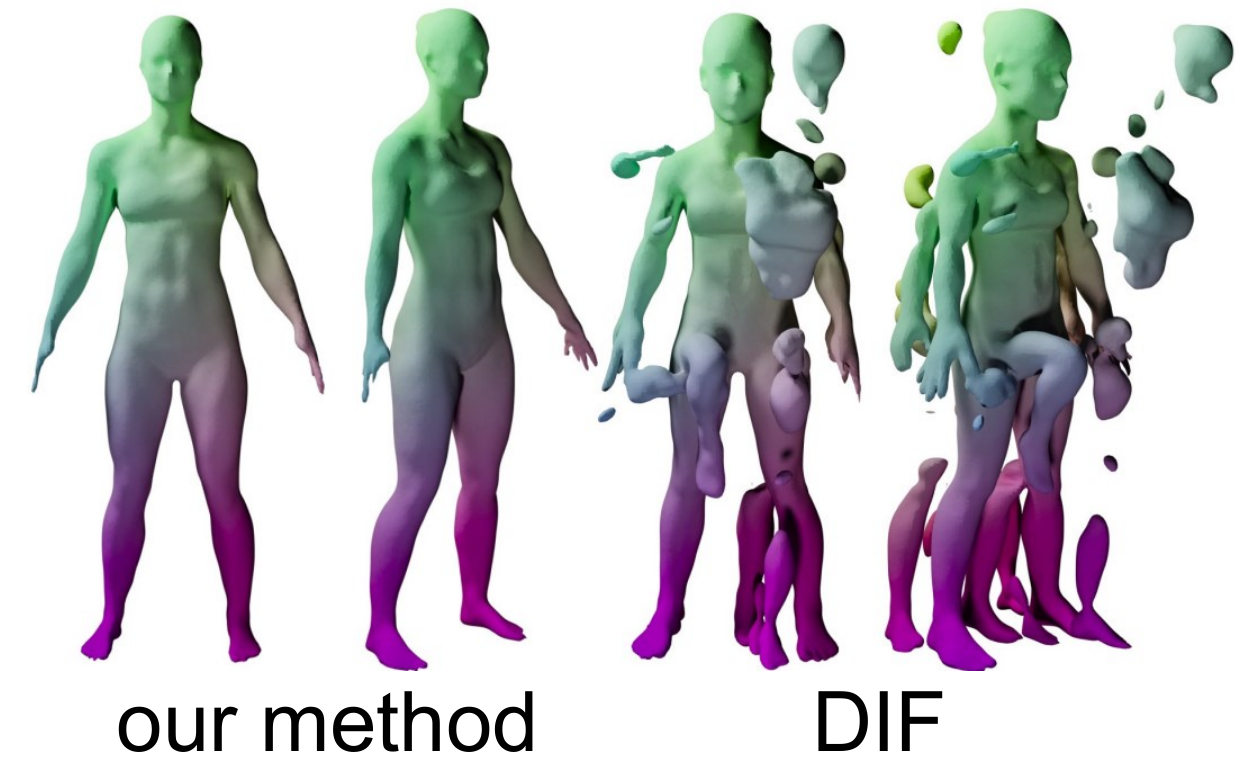}
\caption{Comparison of zero-level set of the learned template field by our method and DIF~\cite{Deng_2021_CVPR}.}
\label{fig:template}
\end{figure}

\subsection{Ablation Study}
We first investigate the effect of our contributed constraints on representation capability.
We conduct comparisons by removing $L_{recon}$, $L_{pr}$, $L_{nbr}$, $L_{pfn}$, $L_{lr}$and $L_{arap}$, respectively, and we also compare the results if  
the $L_{lr}$ only has $L_{arap}$ ('only $L_{arap}$') or the $L_{lr}$ is replaced by the elastic loss \cite{park2021nerfies} ('replace $L_{lr}$').
As shown in Table~\ref{tab:ablation}, all our constraints are useful in our method. Our local rigid constraint is more effective than the elastic loss \cite{park2021nerfies}, and our local rigid constraint could resolve reflection issue of the elastic loss.
Figure~\ref{fig:ablation} shows visual results of our rigid loss terms. Different colors indicate correspondence.
Without $L_{pr}$, the model suffers from heavy artifacts.
The loss $L_{nbr}$ is effective to reconstruct flexible regions such as front legs. 
For 'only $L_{arap}$', 
we do not use the second term of $L_{lr}$ to further penalize the reflection issue on top of $L_{arap}$, and we see inside-out flip at the bear's right front paw.
For 'wo $L_{arap}$', we penalize the second term of $L_{lr}$ (Eq.~\ref{eq:lr}). Figure \ref{fig:wo_arap} shows that this term not only alleviates the floating artifacts, but also helps to find correct correspondences.
So, the term $L_{arap}$ can reduce irregular shapes, and the second term of $L_{lr}$ helps eliminate floating artifacts.
\begin{figure}[ht]
    \centering
    \includegraphics[width=0.5\textwidth]{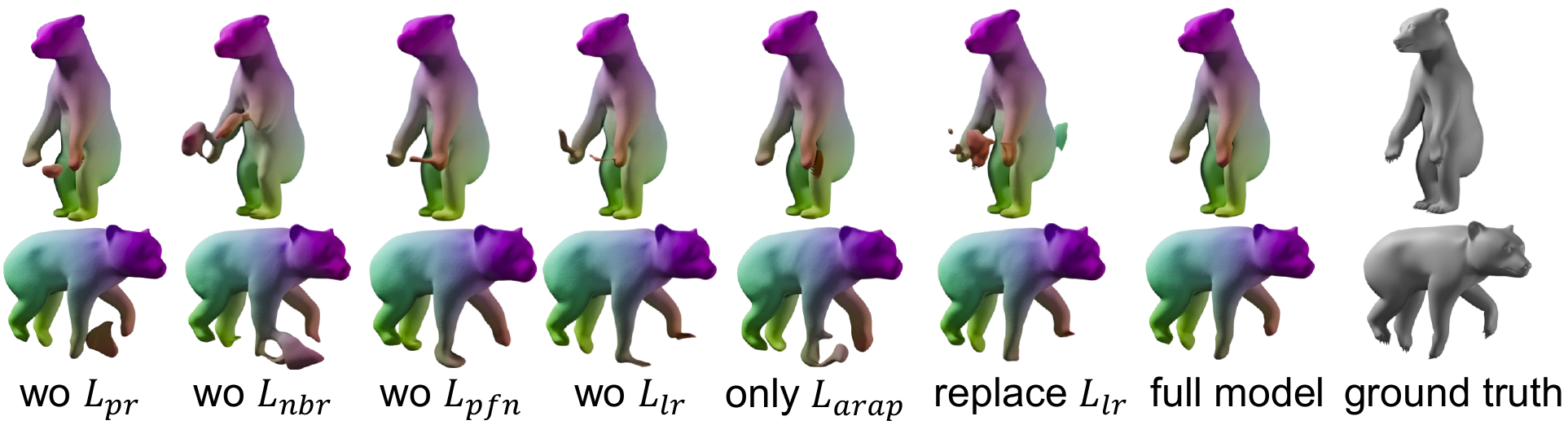}
    \caption{Qualitative experiments of the rigid loss terms.}    
    \label{fig:ablation}
\end{figure}

\begin{figure}[ht]
    \centering
    \includegraphics[width=0.5\textwidth]{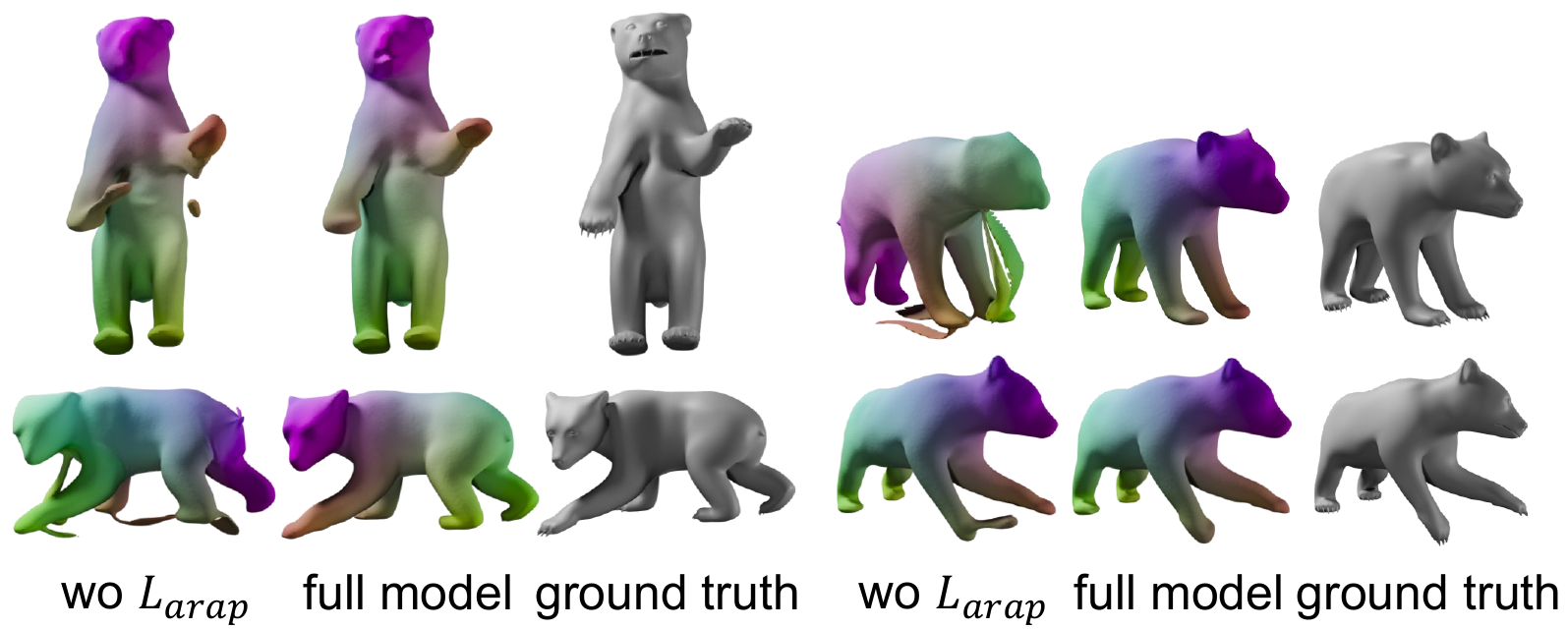}
    \caption{Qualitative experiments of wo $L_{arap}$ terms.}    
    \label{fig:wo_arap}
\end{figure}

\begin{figure}[ht]
\centering
\includegraphics[width=0.5\textwidth]{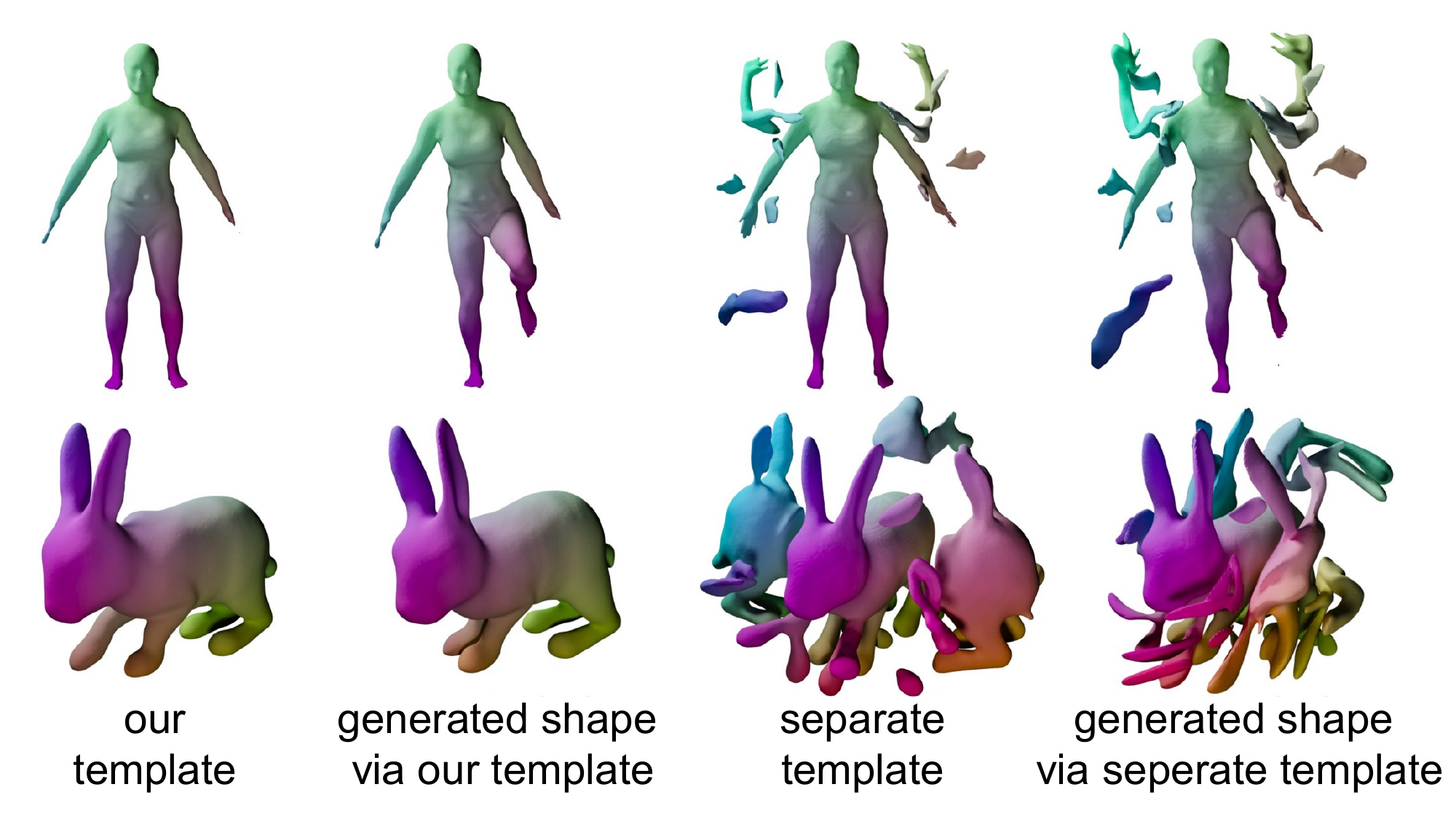}
\caption{Qualitative comparison of our embedded template shape and separate template for shape representation. 
}
\label{fig:template_ablation}
\end{figure}

We also investigate the effect of our embedded template space. We replace 
our embedded template module with a separate network that has the same structure as $\Phi$ to predict template field. 
As shown in Figure~\ref{fig:template_ablation}, the separate template representation learns unreasonable template, and further reduces the ability of our method to represent shapes.

To investigate the effect of different scope of rigid constraints, we change the number of parts in piece-wise rigid constraint, and the standard deviation $\sigma$ of the random sample in Eq.~\ref{eq:neighborhood_rigid} for the neighborhood rigid constraint, and show the quantitative results in Table~\ref{tab:pr_multiscale} and Table~\ref{tab:nbr_multiscale}.
In terms of the physic scale, $L_{lr}$ works in infinitesimal scope. 
The mean volume, proportional to cube of $\sigma$, of convex hull of sampled points in neighborhood constraint is roughly $5\times 10^{-4}$ when $\sigma$ is 0.05, and the mean volume of parts is 0.0768, 0.0334, 0.02175 for part number 5, 10, 20 respectively.
We find that $\sigma=0.05$ corresponds to preferred scale while smaller and lager sale lead to worse performance.
With relatively small scope, piece-wise rigid constraint is flexibly applied on shapes and performs well.

\begin{table*}[ht]
\centering
\resizebox{0.9\linewidth}{!}{
\begin{tabular}{|c|c|c|c|c|c|c|c|c|c|}
\hline
                           & wo $L_{recon}$ & wo $L_{arap}$   & wo $L_{pr}$ & wo $L_{nbr}$ & wo $L_{pfn}$ & wo $L_{lr}$ & only $L_{arap}$ & replace $L_{lr}$ & full model \\ \hline
CD $\downarrow$            & 358.129        & 0.767  & 0.736       & 2.045        & 0.547        & 0.735       & 0.609           & 0.815            & \textbf{0.439}      \\ \hline
\textit{corr $\downarrow$} & 0.0867         & 0.0772 & 0.0858      & 0.0789       & 0.0772       & 0.0709      & 0.0761          & 0.0701           & \textbf{0.0700}     \\ \hline
\end{tabular}
}
\caption{Ablation study on bear from DeformingThings4D~\cite{li20214dcomplete}. We conduct the experiment on training data.}
\label{tab:ablation}
\end{table*}

\begin{table}[h]
\centering
\resizebox{0.7\linewidth}{!}{
\begin{tabular}{|c|c|c|c|}
\hline
                          & 5 parts & 10 parts & 20 parts        \\ \hline
CD$\downarrow$            & 2.169   & 2.645    & \textbf{0.687}  \\ \hline
IoU$\uparrow$             & 0.876   & 0.865    & \textbf{0.885}  \\ \hline
\textit{corr$\downarrow$} & 0.0448  & 0.0375   & \textbf{0.0141} \\ \hline
\end{tabular}
}
\caption{Results of different number of parts in piece-wise rigid constraint.}
\label{tab:pr_multiscale}
\end{table}

\begin{table}[h]
\centering
\resizebox{0.8\linewidth}{!}{
\begin{tabular}{|c|c|c|c|c|}
\hline
                          & $\sigma$=0.01 & $\sigma$=0.03 & $\sigma$=0.1 & $\sigma$=0.05   \\ \hline
CD$\downarrow$            & 1.109         & 1.421         & 1.803        & \textbf{0.687}  \\ \hline
IoU$\uparrow$             & 0.877         & 0.872         & 0.852        & \textbf{0.885}  \\ \hline
\textit{corr$\downarrow$} & 0.0139        & 0.0226        & 0.0520       & \textbf{0.0141} \\ \hline
\end{tabular}
}
\caption{Results of different standard deviations for the neighborhood rigid constraint.}
\label{tab:nbr_multiscale}
\end{table}

\subsection{Applications}\label{sec:application}
\begin{figure}[ht]
\centering
\includegraphics[width=0.5\textwidth]{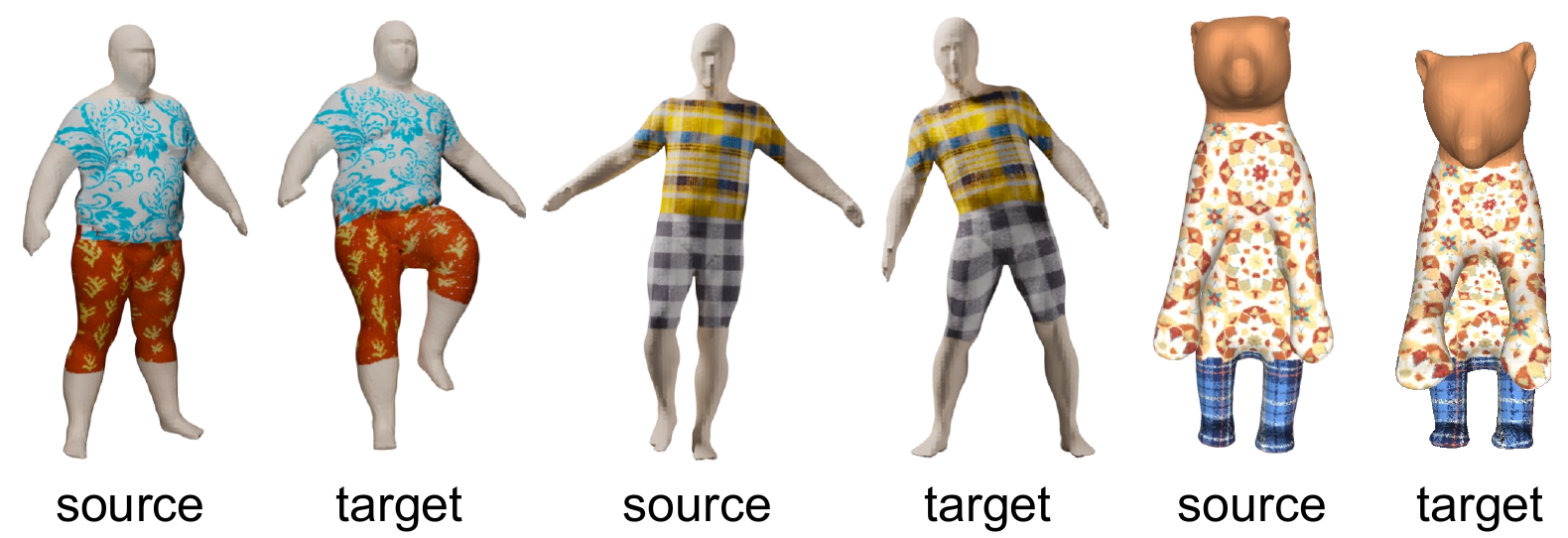}
\caption{Results of texture transfer. Textures on the source shapes are transferred to the target shapes. }
\label{fig:texture}
\end{figure}

\vspace{1mm}
\noindent \textbf{Texture Transfer.} We show a texture transfer application of our method.
Shapes $S_{f_1}$ and $S_{f_2}$ are represented by our model.
We apply texture to $S_{f_1}$, sample points on $S_{f_1}$ and $S_{f_2}$, and transfer these points to the template space. Points on the surface of $S_{f_2}$ query color from the nearest point on the surface of $S_{f_1}$ in template space.
As shown in Figure~\ref{fig:texture}, the textures of the source shapes are well transferred to the correct regions of the target shapes under various poses.  

\vspace{1mm}
\noindent \textbf{Shape Editing.}
We describe how to achieve shape editing with our method. Given a set of surface points ${\mathbf{p}_1}$ on template space and corresponding target points ${\mathbf{p}_2}$, we follow DIF~\cite{Deng_2021_CVPR} to
optimize a latent code $\mathbf{\alpha}_{opt}$ so that the target positions lie on the surface of the generated shape as
{\small
\begin{equation}
\begin{aligned}
    &L=\sum_{\mathbf{p_1},\mathbf{p_2}} w_1 (\big|{\rm \Phi}(\mathbf{p}_2|\alpha_{opt})\big|+\big|{\rm \Phi}(D_{opt\rightarrow tmpl}(\mathbf{p}_2)|\alpha_{tmpl})\big|)\\
    &+w_2 \Vert D_{opt\rightarrow tmpl}(\mathbf{p}_2)-\mathbf{p}_1 \Vert_2^2 
    +w_3 L_{pr}+w_4\Vert\mathbf{\alpha}_{opt}-\mathbf{\alpha}_{tmpl} \Vert_2^2.
\end{aligned}
\label{eq:editing}
\end{equation}
}

\begin{figure}[ht]
\centering
\includegraphics[width=0.45\textwidth]{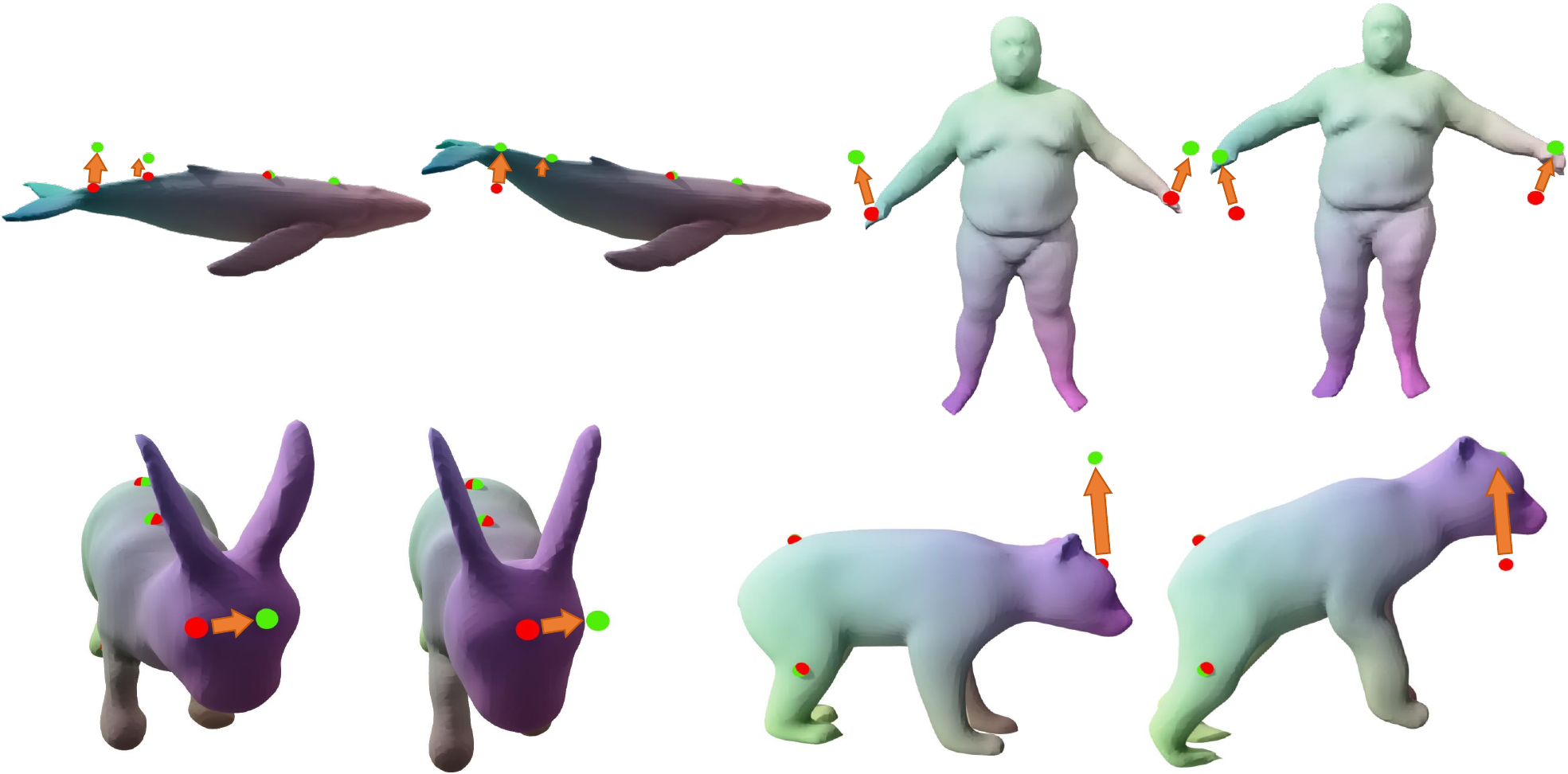}
\caption{Shape editing results. Shapes on the left are warped to shapes on the right. Red points are selected surface points and green points are the target positions. Arrows are used to illustrate the direction of deformations.}
\label{fig:editing}
\end{figure}

The first term constrains the SDFs of target points in target shape and their correspondences in template, the second term enforces correspondence consistency, the third term encourages piece-wise rigid motion, and the fourth term constrains the latent code $\mathbf{\alpha}_{opt}$ close to $\mathbf{\alpha}_{tmpl}$.
To calculate $L_{pr}$, we directly sample points in the bounding box of the template shape and the target points $\mathbf{p}_2$.
As shown in Figure~\ref{fig:editing}, the template shapes are well deformed to target points.
At the top right of the figure, raising the hands can cause the left leg to move in an undesired manner, and the resulting average shape after shape editing may become unreasonable. This issue is out of the scope of this paper and we leave it as the future work.

\section{Limitations and Future Works}
    Although our method can deal with shapes with large deformation and achieve promising results, there are some problems left for future works. 
First, our method cannot represent out-of-distribution data.
 our method acquires knowledge about shape deformation from training set and does not use any prior of skeleton, so it cannot represent shapes that are out of distribution of training set.
This problem limits the generalization of shape reconstruction and degrades shape editing (Sec.~\ref{sec:application}).
Second, our method cannot represent shape collections with topology changes because 
the key features (such as SDF values and normals) used in our method cannot be correctly calculated under this data setting. 
For example, if two body parts contact with each other, we can not get accurate SDF values due to the invisible contact surface from any camera view. 
In fact, this is a common issue for other methods using implicit representation. 
Third, the latent code space lacks smooth constraint. Our method is capable of optimizing the latent code to get a valid interpolation, but cannot guarantee valid shapes when receiving random codes. In fact, most codes in the latent space is invalid, which may be tackled by proper smooth constraint in latent.
In the future work, we plan to address these challenges and propose effective shape representation approaches for more general deformable shapes.

\section{Conclusions}
    We present a model to represent shape with dense correspondence in a self-supervised manner. Even for subjects with large deformation, our method can learn good shape and dense correspondences. 
We also show two typical applications of shape representation, and our method can achieve competitive performance.

\noindent \textbf{Acknowledgments.} This work was supported in part by National Key R\&D Program of China (2022ZD0117900).


{\small
\bibliographystyle{ieee_fullname}
\bibliography{egbib}
}

\setcounter{section}{0}
\renewcommand\thesection{\Alph{section}}
\setcounter{table}{0}
\renewcommand\thetable{\Alph{table}}
\setcounter{figure}{0}
\renewcommand\thefigure{\Alph{figure}}
\setcounter{equation}{0}

\title{Supplementary Material for\\ ``Self-supervised Learning of Implicit Shape Representation\\ with Dense Correspondence for Deformable Objects''}

\author{
Baowen Zhang$^{1,2}$\qquad Jiahe Li$^{1,2}$\qquad Xiaoming Deng$^{1,2}$\thanks{indicates corresponding author}\qquad Yinda Zhang$^3$$^*$ \\   Cuixia Ma$^{1,2}$\qquad Hongan Wang$^{1,2}$\\
$^1$Institute of Software, Chinese Academy of Sciences \quad $^2$University of Chinese Academy of Sciences\\  $^3$Google \quad  
}

\maketitle
\ificcvfinal\thispagestyle{empty}\fi

In this supplementary material, we first introduce the details on network architecture (Section~\ref{sec:network}) and loss functions (Section~\ref{sec:loss}). Then we show additional experiments and ablation study (Section~\ref{sec:exp}). Finally, we give the closed-form analytical solution to get the minimal alignment error of our piece-wise rigid constraint (Section~\ref{sec:closeform}), analyze the least square solution of rotation (Section~\ref{sec:leastrotation}), and reveal the relationship between ARAP loss and the local rigid constraint (Section~\ref{sec:arap}).

\section{Details on Network Architecture}
\label{sec:network}
In this section, we describe more details of our network.
It consists of three modules: an encoder-decoder network, part prediction networks, and an SDF prediction module.

\vspace{1mm}
\noindent \textbf{Encoder-Decoder Network.} The encoder-decoder network predicts correspondences on template (Figure~2 of the main paper).
The encoder receives a point $\mathbf{p}$ from the target space $S_i$, along with its predicted SDF value from the SDF prediction module as input. It then produces a vector $\mathbf{l}(\mathbf{p}) \in \mathbb{R}^{8}$.
Following that, the decoder takes $\mathbf{l}(\mathbf{p})$ as input and outputs the corresponding point $D_{i\rightarrow tmpl}(\mathbf{p})$ in template space.
Our encoder and decoder consist of 5 and 4 fully connected layers, respectively.

\vspace{1mm}
\noindent \textbf{Part Prediction Networks.} The networks $\psi_e$ and $\psi_d$ predict part probabilities, \ie $\mathbf{P}_h$ in Eq.~6 of the main paper, for each point.
Each network divides the target shape into 20 parts, totally 40 parts together for calculating piece-wise rigid constraint, \ie, $N_P=40$ in Eq.~6 of the main paper.
Our part prediction networks $\psi_e$ and $\psi_d$ have 4 and 3 fully connected layers, respectively. 
We use SoftMax to normalize the probabilities predicted by each network.
Both networks are trained in a self-supervised manner by optimizing piece-wise rigid constraint.

For each point $\mathbf{p}$ in a target space $S_i$, $\psi_e$ takes the output vector $\mathbf{l}(\mathbf{p}) \in \mathbb{R}^{8}$ of the encoder as input, and $\psi_d$ takes the correspondence point $D_{i\rightarrow tmpl}(\mathbf{p})\in S_{tmpl}$ in template space as input.
Since the correspondences are consistent across shapes deformed from the same template, the part segmentation learned by $\psi_d$ is also consistent across all shapes.
The segmentation results are shown in Figure~4 in our main paper and Figure~\ref{fig:seg} in the supplementary material.
However, the correspondences are not learned well at the beginning of training.
Conceptually, the prediction of $\psi_d$ highly depends on learned correspondences and template,
so it cannot be effectively trained at the beginning of training, with highly-undertrained optimization of correspondences and template.
In order to address this issue, we use $\psi_e$ to predict part segmentation, which does not depend on correspondences on template.
During training stage, we observe that $\psi_e$ provides valid rigid constraint earlier than $\psi_d$ and enables the network to converge faster.

\vspace{1mm}
\noindent \textbf{SDF Prediction Module.} The SDF prediction module ${\rm \Phi}$ is in charge of modeling template SDF field as well as SDF fields of other shapes in training set. With dense correspondence predicted by the encoder-decoder network, we can query SDF values from template field to reconstruct target shapes (Eq.~2 in the main paper). Inspired by Atzmon \etal \cite{atzmon2020sal} that the initial scalar field contributes greatly to shape representation learning, we add the distance of an input point $\mathbf{p}$ to center $(0,0,0)$ to the output of the neural implicit SDF function ${\rm \Phi}$ and achieve similar initialization to Atzmon \etal \cite{atzmon2020sal}.
The output of SDF prediction module is formulated like \cite{IROS_2022_ReDSDF} as ${\rm \Phi}(\mathbf{p}|\mathbf{\alpha})=\phi(\mathbf{p}|\mathbf{\alpha})+\Vert \mathbf{p} \Vert_2$, where $\phi$ denotes a neural network for SDF prediction. The network $\phi$ consists of 5 fully connected layers.

\vspace{1mm}
\noindent \textbf{Other Details.} Similar to the previous works~\cite{Deng_2021_CVPR,sitzmann2019scene}, the parameters of encoder, decoder, and SDF prediction module are all predicted by Hyper-Nets, while part probability networks have their own parameters.
All the Hyper-Nets in our method consist of 5 fully connected layers with $relu$ as activation function.
The dimension of hidden features is 256 in Hyper-Nets, and is 128 in other modules. The dimension of latent code ${\rm \alpha}$ for each shape is 128.

We use the sine activation function proposed by Sitzmann \etal \cite{sitzmann2020implicit} for encoder, decoder, SDF module and part probability networks, because it has excellent property of representing complex signal and its derivative \cite{sitzmann2020implicit}.
The sine activation function is in form of $f(\mathbf{x})=sin(\omega \mathbf{x})$, and larger $\omega$ usually indicates output with higher frequency.
In our experiments, $\omega$ is set to 15 in part probability networks, and set to 30 in encoder, decoder and SDF prediction networks.

\section{More Details on Loss Functions}
\label{sec:loss}
In Section~3.4 of our main paper, we follow the idea of $L_{sdf}$ to supervise queried SDF values from template.
In the following, we show the detailed formulations of the constraints.
The $L_{sdf}$ is used to supervise SDF values ${\rm \Phi}(\mathbf{p}|\mathbf{\alpha_i})$, while the following constraints are used to supervise SDF values queried from template field ${\rm \Phi}(D_{i\rightarrow tmpl}(\mathbf{p})|\mathbf{\alpha}_{tmpl})$.

\vspace{1mm}
\noindent \textbf{SDF Regression Constraints.}  
The SDF regression constraints have the similar formulation to $L_{sdf}$.
In order to constrain the queried SDF to have the same sign of the ground truth, we design the constraint for queried SDF value formulated as
\begin{equation}
\begin{aligned}
    &L_{pbs}=\sum_{\mathbf{p}\in S_i} |\hat{s}(\mathbf{p})|,\\
    \hat{s}(\mathbf{p})=&\resizebox{.45\textwidth}{!}{$\left\{
    \begin{aligned}
    &{\rm \Phi}(D_{i\rightarrow tmpl}(\mathbf{p})|\alpha_{tmpl}),\ {\rm if}\ \bar{s}\cdot {\rm \Phi}(D_{i\rightarrow tmpl}(\mathbf{p})|\alpha_{tmpl})\leq0 \\
    &0,\ otherwise
    \end{aligned}
    \right.$}
\end{aligned}
\end{equation}
where $\bar{s}$ is the ground truth SDF value. The loss weight of $L_{pbs}$ is $w_s$, which  is the same as $w_s$ in the main paper (the first term of Eq.~8).

In order to supervise the normal on a represented shape, we constrain the gradient of the queried SDF field to align with ground truth normal:
\begin{equation}
    L_{pbn} = \sum_i \sum_{\mathbf{p}\in S_i^0} (1-S_c(\nabla_\mathbf{p} {\rm \Phi}(D_{i\rightarrow tmpl}(\mathbf{p})|\alpha_{tmpl}),\bar{\mathbf{n}})),
\end{equation}
where $\bar{\mathbf{n}}$ is the ground truth normal, and $S_c$ is cosine similarity.
Note that the $L_{pbn}$ is different from $L_{pfn}$ (Eq.~9 in the main paper).
The loss $L_{pbn}$ supervises normals on the represented shape, while $L_{pfn}$ supervises normals on template, which is proved to be crucial for shape representation learning by Deng \etal \cite{Deng_2021_CVPR}.
The weight of $L_{pbn}$ and $L_{pfn}$ is $w_n$ (which is also the weight of $\sum_{\mathbf{p}\in S_i^0}(1-S_c(\nabla\Phi(\mathbf{p}|\alpha_i),\bar{\mathbf{n}})$ of Eq.~8 in the main paper).

We also constrain the gradient of template field to satisfy Eikonal equation:
$\sum_{\mathbf{p}\in{S_{tmpl}}}\vert \Vert \nabla{\rm \Phi}(\mathbf{p}|\alpha_{tmpl}) \Vert_2-1 \vert$, and apply $\rho$ (the fourth term of Eq.~8 of the main paper) on template field to encourage off-surface points on template to have larger SDF values. Their weights are same as $w_{Eik}$ and $w_{\rho}$ in the main paper.

\vspace{1mm}
\noindent \textbf{Reconstruction Loss.} 
We give the detailed formulation of $L_{recon}$ in Section~3.4 of the main paper as

\begin{equation}
\begin{aligned}
    L_{recon}=&\sum_i\sum_{\mathbf{p}\in S_i}\Vert \mathbf{p}-D_{i\rightarrow i}(\mathbf{p})\Vert^2+\\
    &\sum_{\mathbf{p}\in S_{tmpl}}\Vert \mathbf{p}-D_{tmpl\rightarrow tmpl}(\mathbf{p})\Vert^2.
\end{aligned}
\end{equation}
The weight of each loss term remains the same across all subjects, specially $w_s=3\times 10^2$, $w_n=50$, $w_{Eik}=5$, $w_{\rho}=50$, $w_{recon}=5\times 10^3$, $w_{reg}=1\times 10^5$, $w_{lr}=10$, $w_{nbr}=5\times 10^4$, $w_{pr}=3\times 10^3$.

\begin{figure}[h]
    \centering
    \includegraphics[width=0.5\textwidth]{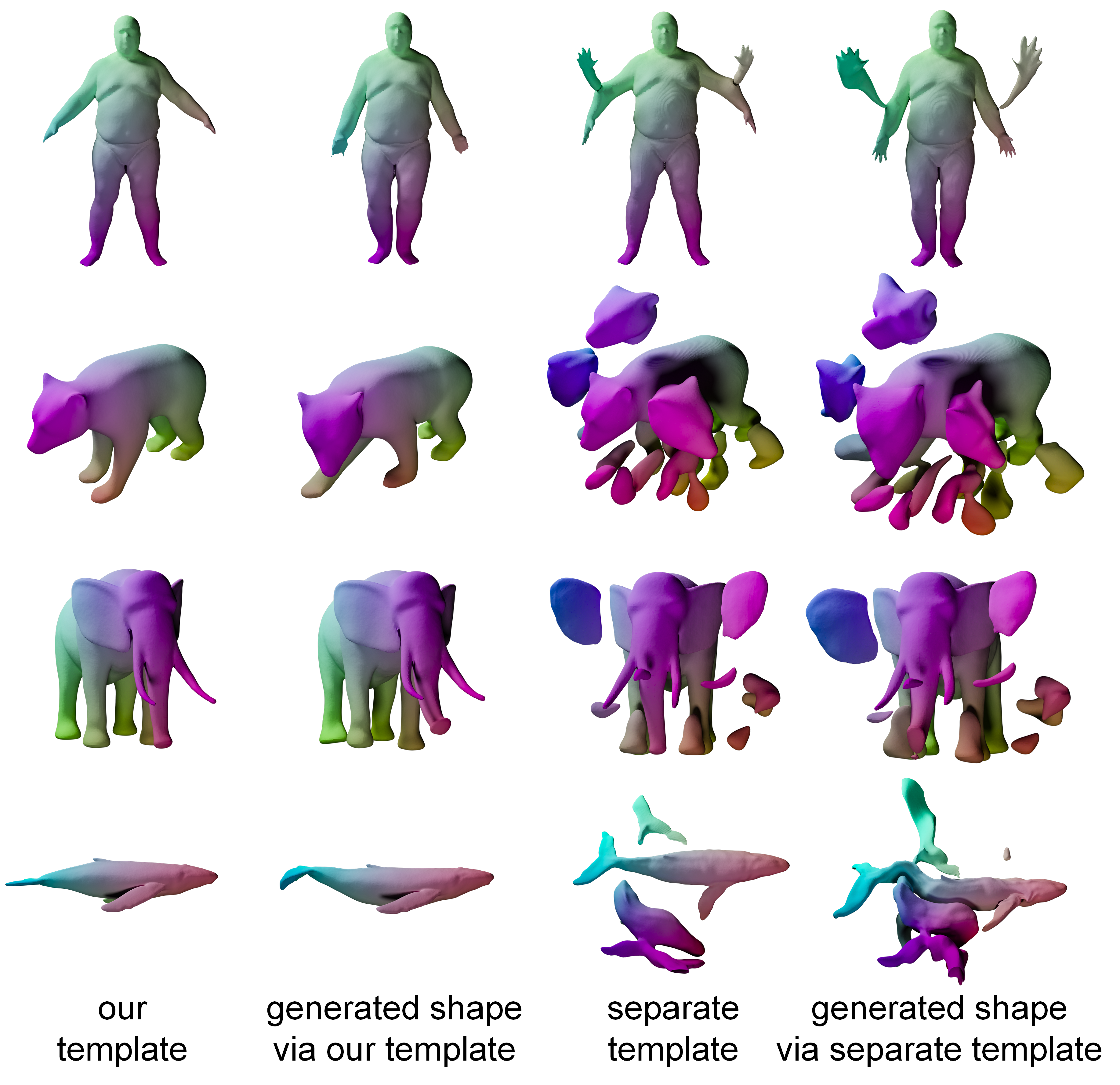}
     \caption{More qualitative experiments to demonstrate the ability of our method to learn template. 
    The learned template shapes with the separate template representation are all not reasonable, and our template enables significantly better results of generated shapes than those with the separate template. 
     }
    \label{fig:template}
\end{figure}
\section{Additional Experiments}
\label{sec:exp}
\begin{table}
\centering
\begin{tabular}{|c|c|c|c|}
\hline
 &w/o $\psi_e$&w/o SDF input&full model\\ \hline
CD $\downarrow$ &1.174 &0.783&\textbf{0.687}\\ \hline
\emph{corr} $\downarrow$&0.0165 &0.0265 &\textbf{0.0141}\\ \hline
\end{tabular}
\caption{Ablation study on subject 50026 from D-FAUST\cite{dfaust:CVPR:2017} dataset. "w/o $\psi_e$" represents the model without part probabilities prediction network $\psi_e$. "w/o SDF input" represents the input of our encoder only contains coordinate $\mathbf{p}$. It shows that $\psi_e$ can improve the performance on both Chamfer distance and geodesic distance. Using SDF as input of encoder can greatly improve the performance of our method.}
\label{tab:ablation}
\end{table}

\subsection{Ablation Study}

\vspace{1mm}
\noindent \textbf{Effect of Our Template Representation.}
In this section, we use the same method as the main paper to further investigate the ability of our novel template shape representation architecture and show more results.
Instead of representing the template shape as a latent code, we test an ablation case where a separate network only predicts template SDF like DIF.
The architecture of the new template SDF module is the same as the origin SDF prediction module in our main paper.
Other parts of our network and the loss functions remain the same.
More results are shown in  Figure~\ref{fig:template}. The learned template shapes with the separate template representation are all not reasonable, and our template enables significantly better results of generated shapes than those with the separate template. 

\begin{figure}[t]
    \centering
    \includegraphics[width=0.5\textwidth]{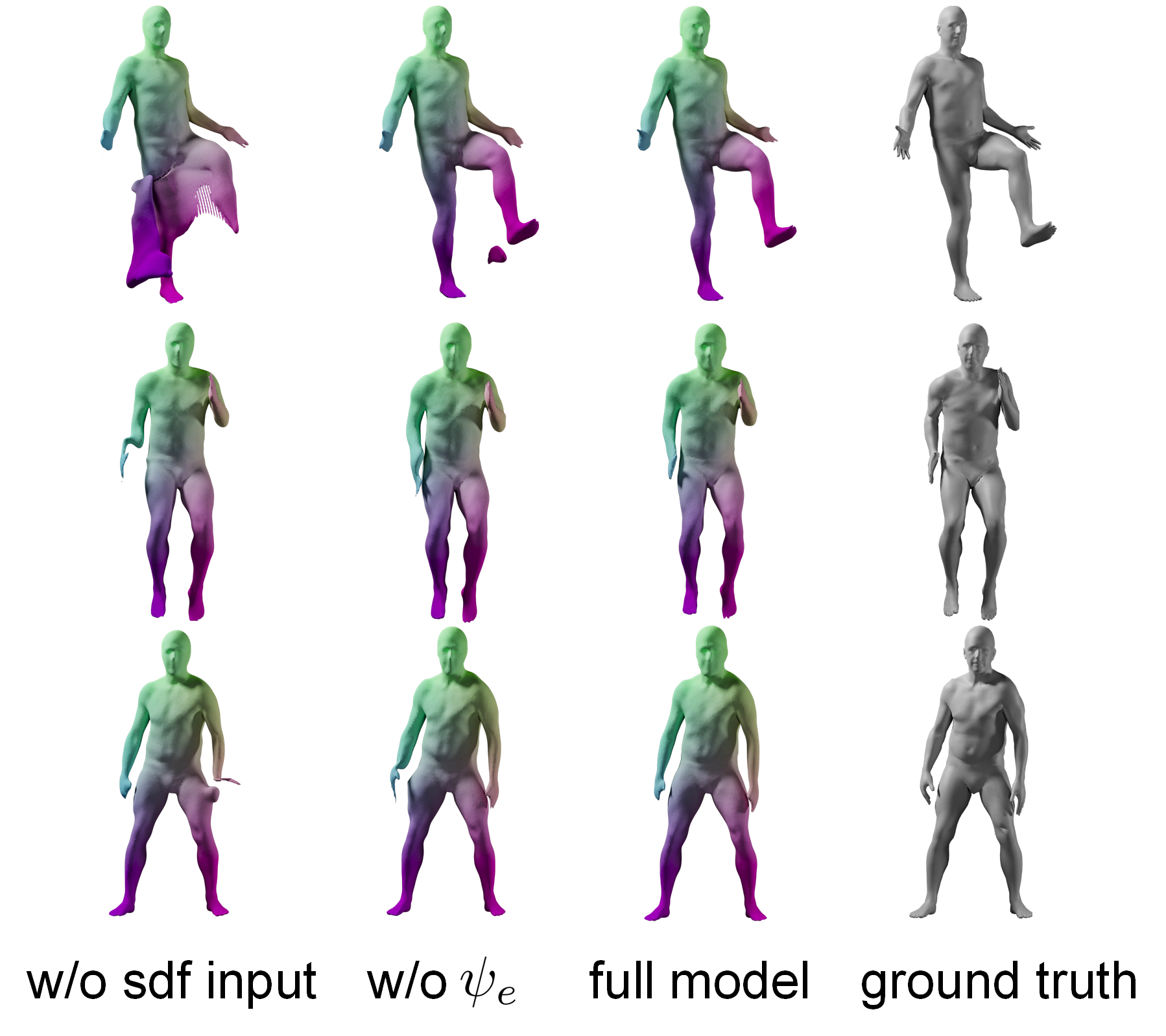}
    \caption{Qualitative results of ablation study. Colors indicate the dense correspondences. SDF can provide encoder with geometric clues to predict correspondence. The part prediction network $\psi_e$ is an essential component of the rigid constraint, which helps the network converge.}
    \label{fig:ablation}
\end{figure}

\vspace{1mm}
\noindent \textbf{Effect of $\psi_e$ in Piece-wise Rigid Constraint.}
In the main paper, we conduct the experiments to demonstrate the significance of piece-wise rigid constraint.
In this ablation, we investigate the effect of the part probabilities  predicted by the networks $\psi_e$ (See Section~\ref{sec:network}).
Although we cannot attain consistent part segmentation across shapes with $\psi_e$ only, we demonstrate $\psi_e$ plays a crucial role in our method.
We compare the reconstruction and correspondence results of our method without $\psi_e$ on subject 50026 in D-FAUST~\cite{dfaust:CVPR:2017} in Table~\ref{tab:ablation} and Figure~\ref{fig:ablation}. Subject 50026 is selected for evaluation due to its large range of motion. Experiments show that $\psi_e$ can deal with this challenge and improve the performance of shape reconstruction and correspondence prediction. In contrast, we observe that method without $\psi_e$ cannot converge as well as our full model, especially in the end points of the body with the large range of motion.

\vspace{1mm}
\noindent \textbf{Effect of SDF input to Encoder.} In order to evaluate the effect of SDF input to encoder, we compare the Chamfer distance of shape reconstruction and geodesic distance of predicted correspondences by removing the SDF input. We also show the results using our method without input SDF on subject 50026 in D-FAUST~\cite{dfaust:CVPR:2017} in Table~\ref{tab:ablation} and Figure~\ref{fig:ablation}. We observe that method without SDF as input to encoder fails in some poses and generates shapes with bad geometry.

\subsection{More Evaluations on Model Capacity}
In this section, we show more model capacity comparisons with DIF~\cite{Deng_2021_CVPR} and 3D-CODED~\cite{groueix2018b} using reconstructed shapes in the training set. 
Figure~\ref{fig:comptraininghuman} and Figure~\ref{fig:comptraininganimal} shows that our method outperforms both of them.

\subsection{Reconstruction from Full Observation}
In this section, we will show more qualitative  experiments of our method compared with DIF~\cite{Deng_2021_CVPR} and 3D-CODED~\cite{groueix2018b}.
We generate point cloud of each subject by simulating multiple depth cameras,
and then fit our shape representation model by minimizing Eq.~10 in the main paper.
Figure~\ref{fig:compfullhuman} and Figure~\ref{fig:compfullanimal} show the results of humans and animals.
We can observe that our method outperforms DIF and 3D-CODED, and achieves plausible shape reconstruction and correspondence results.
Our method can fit shapes with large deformation effectively.
Conceptually, 3D-CODED and DIF lack sufficient rigid constraints, so they cannot model subjects with large deformation properly.
Although DIF can learn template SDF field, the learned shape is out of the distribution of the training data.
Therefore, there are many floating components on the reconstructed shapes.

\subsection{Reconstruction from Partial Observation}
We generate point cloud of each subject by simulating a single depth camera,
and then fit our representation model by minimizing Eq.~10 in the main paper.
Figure~\ref{fig:comppartialhuman} and Figure~\ref{fig:comppartialanimal} show the qualitative experiments of shape reconstruction from partial point cloud.
Our model can reconstruct shapes from partial point cloud while 3D-CODED~\cite{groueix2018b} fails. Therefore, we only compare with DIF~\cite{Deng_2021_CVPR}. Results show that our method
outperforms DIF by a large margin for partial point clouds.

\subsection{Comparison with LoopReg}
In this section, we compare our method with LoopReg \cite{bhatnagar2020loopreg} on training set. 
LoopReg creates a self-supervised loop to register a corpus of scans to a common 3D human model (\ie, SMPL \cite{loper2015smpl}), which can model correspondences between human pairs.
As shown in Table~\ref{tab:compLoopReg}, our method outperforms LoopReg on both IoU and \emph{corr}.
\begin{table}[h]
\centering
\begin{tabular}{|c|c|c|}
\hline
     & LoopReg & Our method   \\ \hline
IoU $\uparrow$  & 0.726   & \textbf{0.881}  \\ \hline
\emph{corr} $\downarrow$ & 0.1087   & \textbf{0.0304} \\ \hline
\end{tabular}
\caption{Capacity evaluation on D-FAUST with LoopReg \cite{bhatnagar2020loopreg} }
\label{tab:compLoopReg}
\end{table}

\subsection{Qualitative Experiment on Part Segmentation}
\begin{figure}[t]
    \centering
    \includegraphics[width=0.5\textwidth]{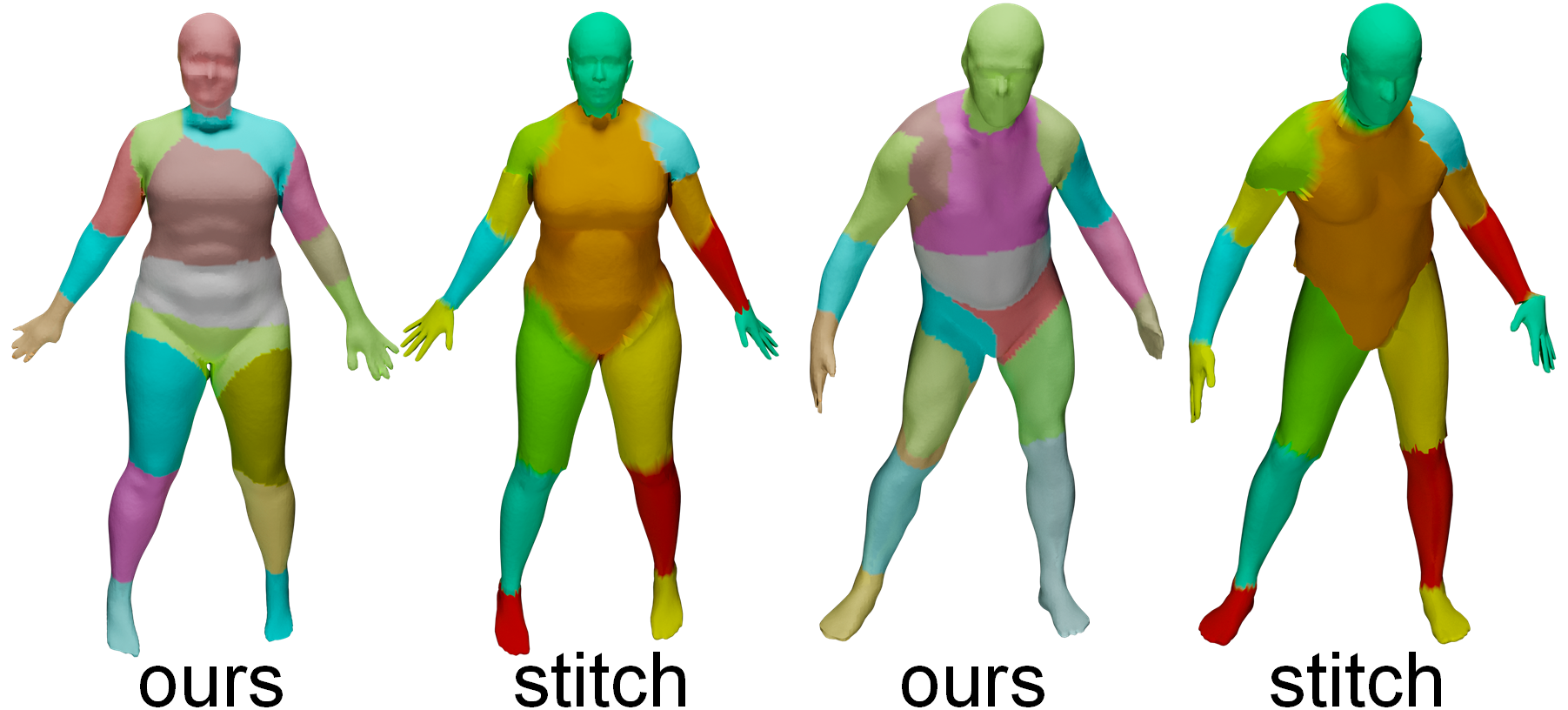}
    \caption{Comparison with the stitched puppet \cite{zuffi2015stitched}.}
    \label{fig:seg}
\end{figure}

In this section, we compare our method with the stitched puppet \cite{zuffi2015stitched}.
The stitched puppet \cite{zuffi2015stitched} is a shape representation method that manually segments the represented shape into multiple parts and combines the parts into human body shapes with different poses.
As shown in Figure~\ref{fig:seg}, our self-supervised method achieves comparable results.
\subsection{Failure Cases}
\begin{figure}[t]
    \centering
    \includegraphics[width=0.5\textwidth]{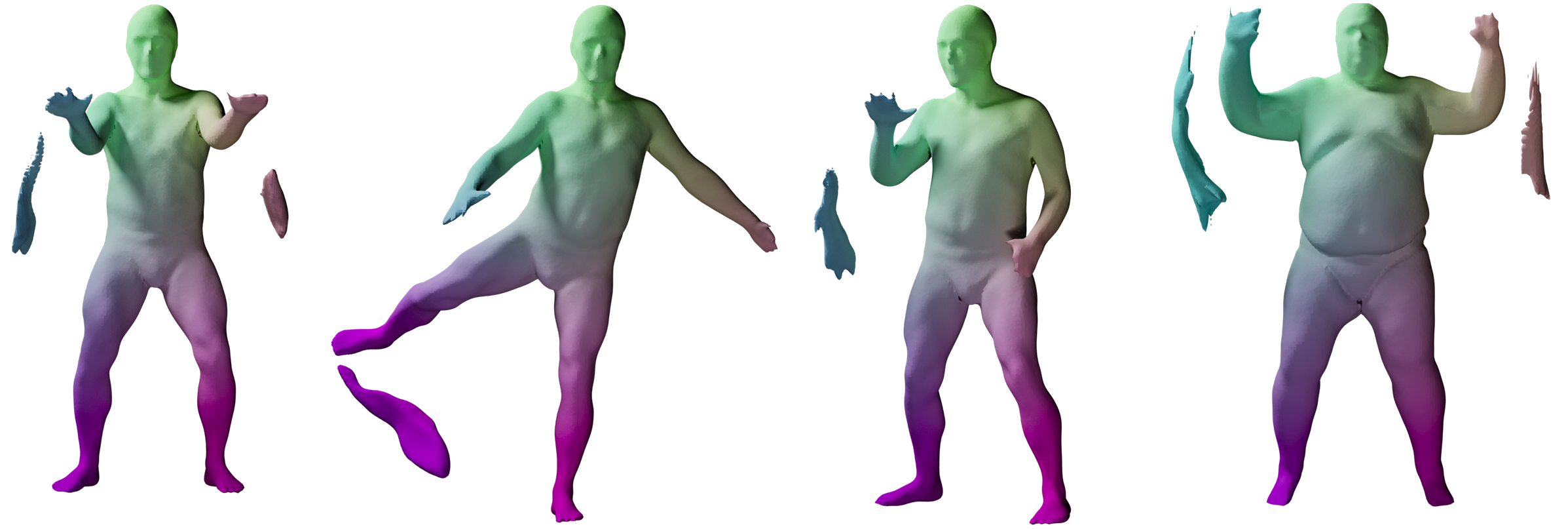}
    \caption{Several failure cases where floating components make Chamfer distance increase.}
    \label{fig:flyer}
\end{figure}

\begin{figure*}[b]
    \centering
    \includegraphics[width=1.0\textwidth]{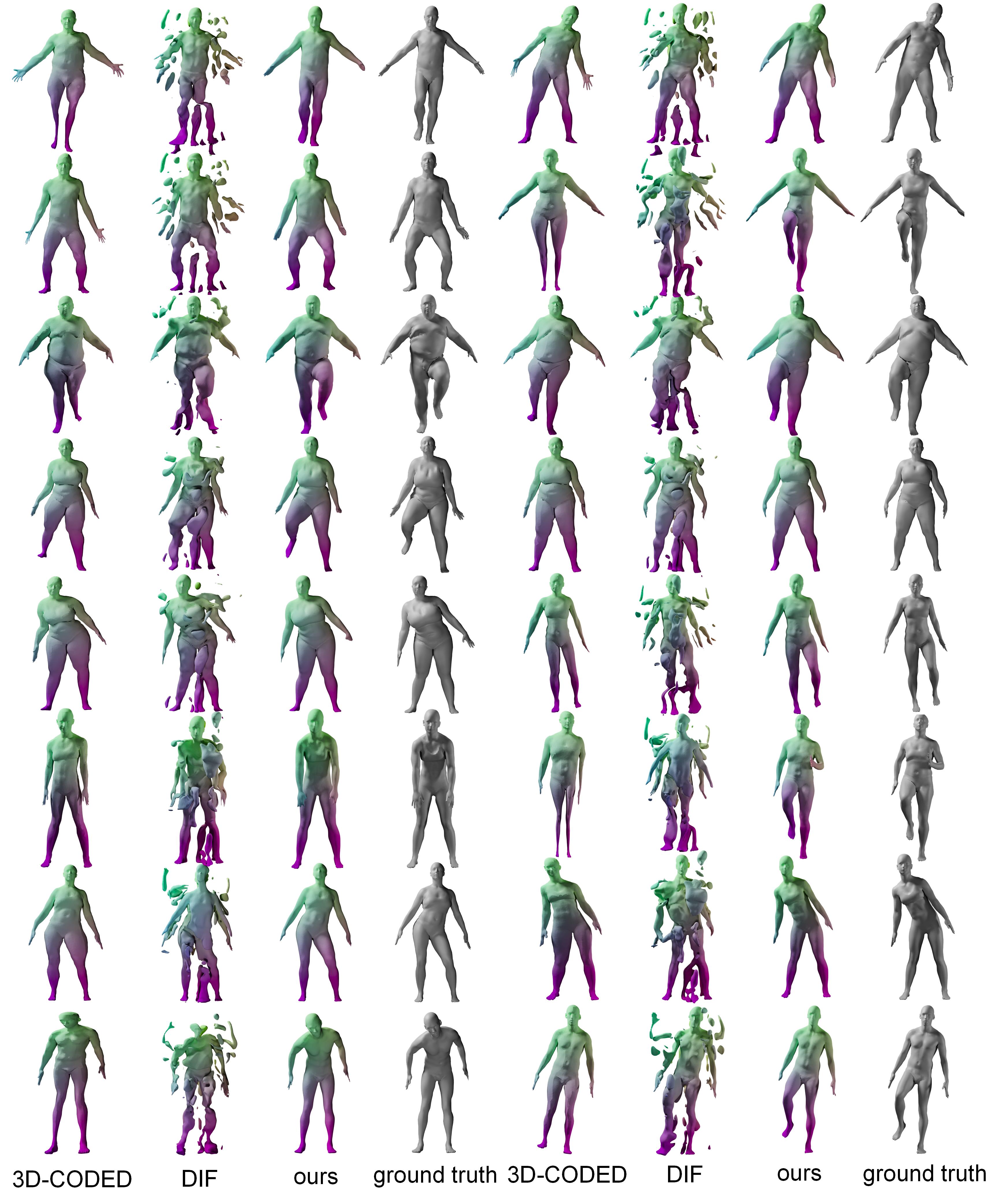}
    \caption{Reconstruction from training set of humans. We compare our method with DIF~\cite{Deng_2021_CVPR} and 3D-CODED~\cite{groueix2018b}. Our method reconstructs shapes with multiple poses and large deformations. The characteristics of each subject is represented well.}
    \label{fig:comptraininghuman}
\end{figure*}

\begin{figure*}[b]
    \centering
    \includegraphics[width=1.0\textwidth]{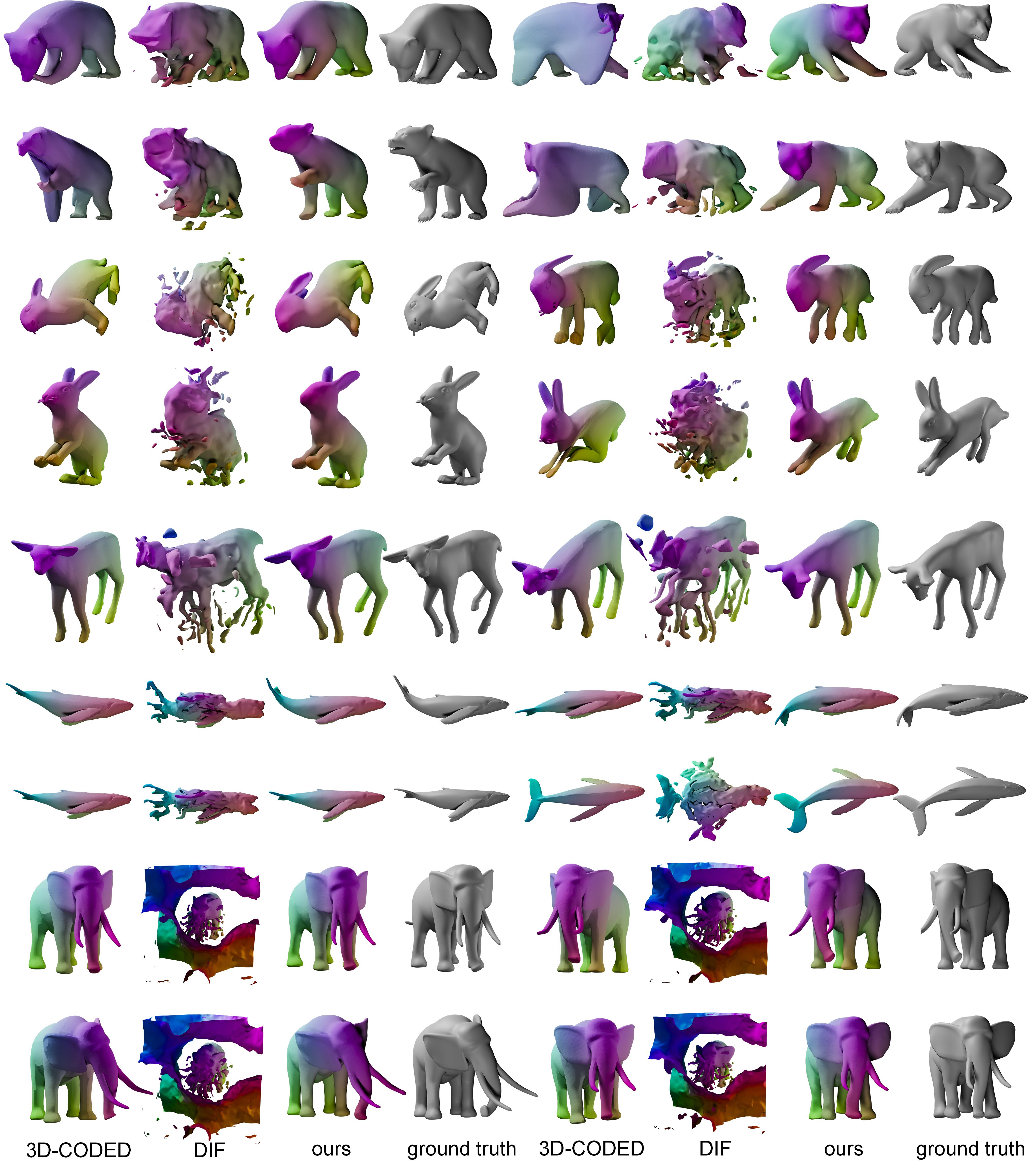}
    \caption{Reconstruction from training set of animals. We compare our method with DIF~\cite{Deng_2021_CVPR} and 3D-CODED~\cite{groueix2018b}. Our method reconstructs shapes with multiple poses and large deformations. The characteristics of each subject is represented well.}
    \label{fig:comptraininganimal}
\end{figure*}

\begin{figure*}[b]
    \centering
    \includegraphics[width=1.0\textwidth]{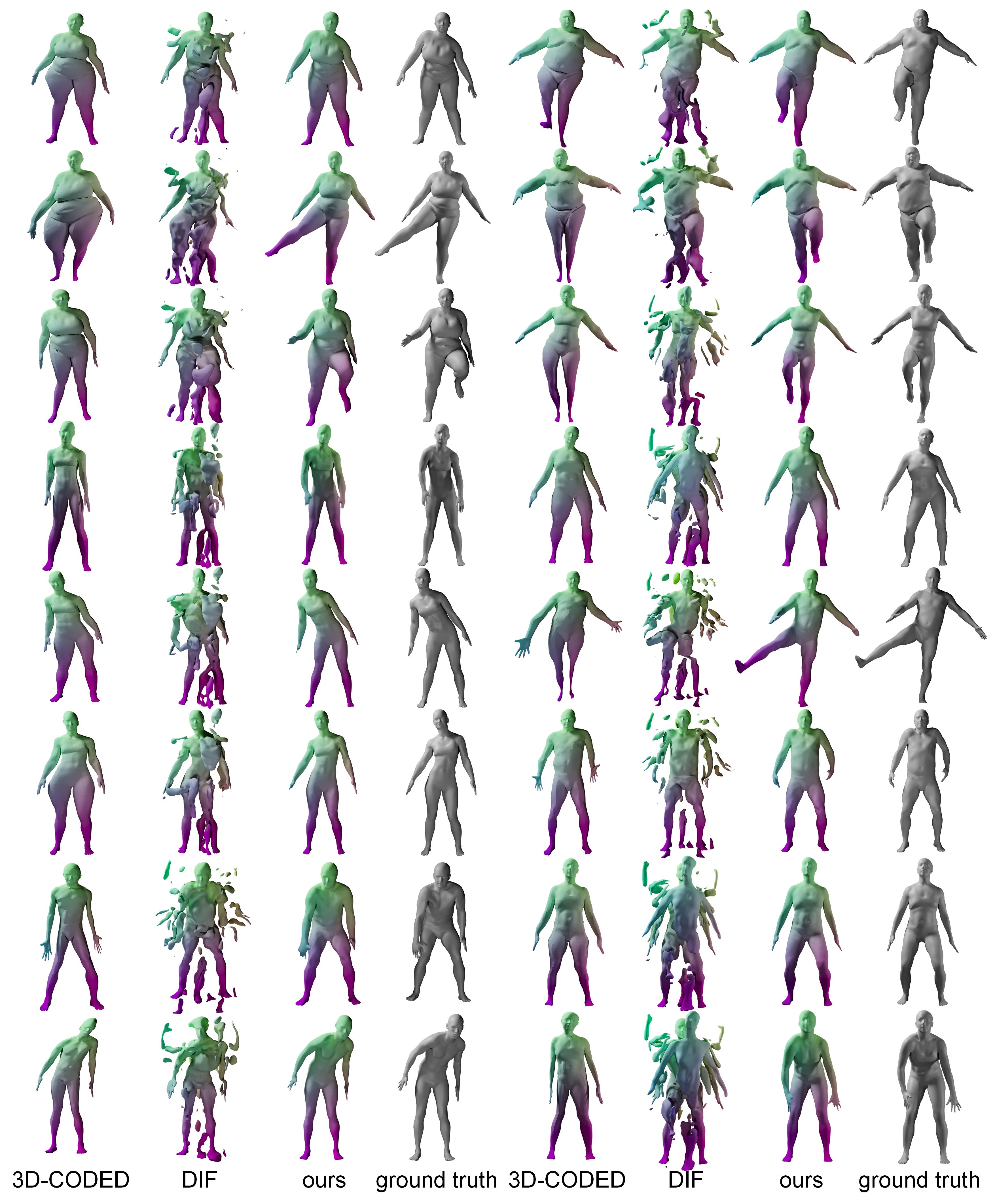}
    \caption{Reconstruction from full observation of humans. We compare our method with DIF~\cite{Deng_2021_CVPR} and 3D-CODED~\cite{groueix2018b}.
    Our method achieves plausible shape reconstructions and can predict correspondence across shapes.
    }
    \label{fig:compfullhuman}
\end{figure*}

\begin{figure*}[b]
    \centering
    \includegraphics[width=1.0\textwidth]{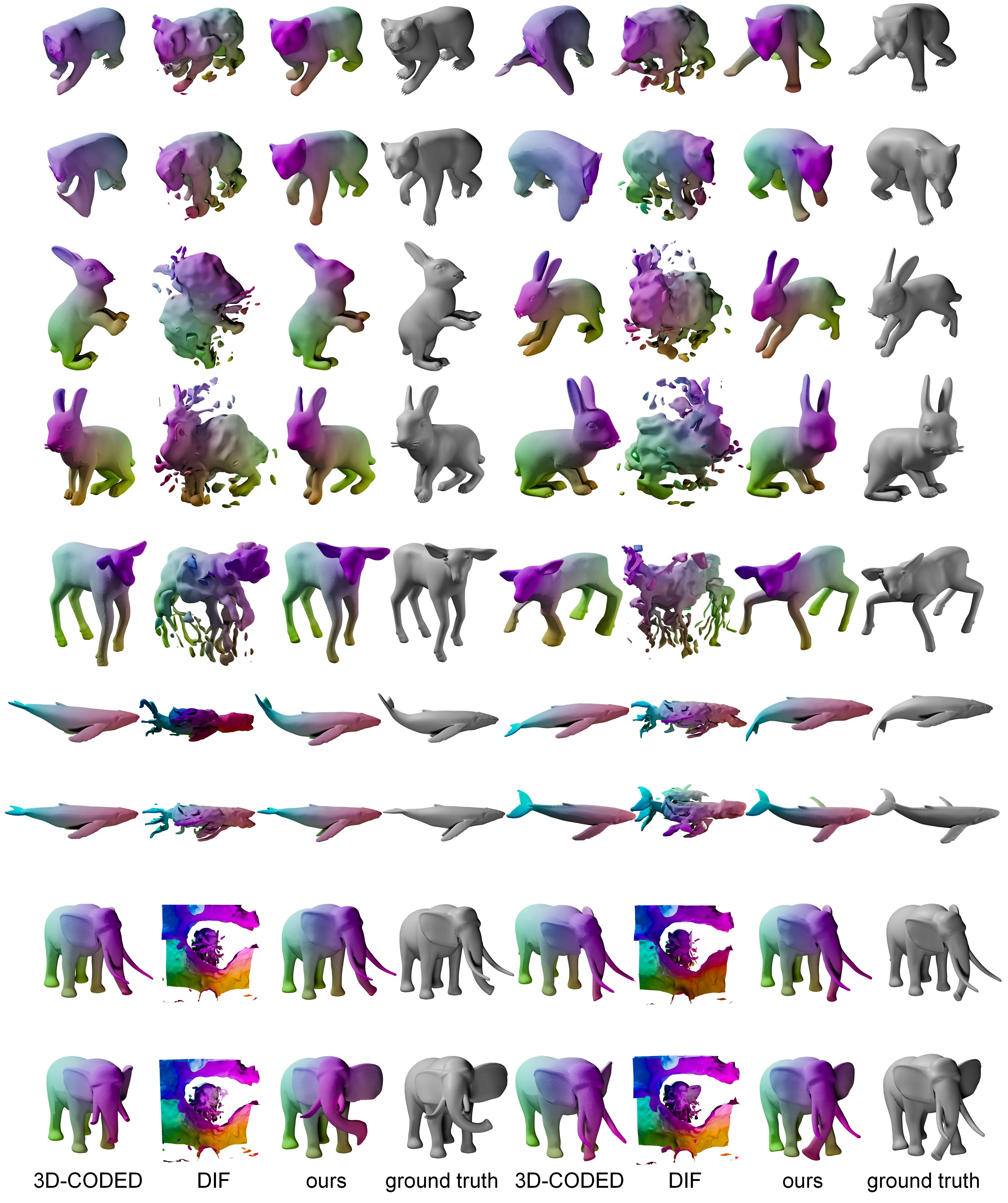}
    \caption{Reconstruction from full observation of animals. We compare our method with DIF~\cite{Deng_2021_CVPR} and 3D-CODED~\cite{groueix2018b}.
    Our method achieves plausible shape reconstructions and can predict reliable correspondence across shapes.
    }
    \label{fig:compfullanimal}
\end{figure*}

\begin{figure*}[b]
    \centering
    \includegraphics[width=1.0\textwidth]{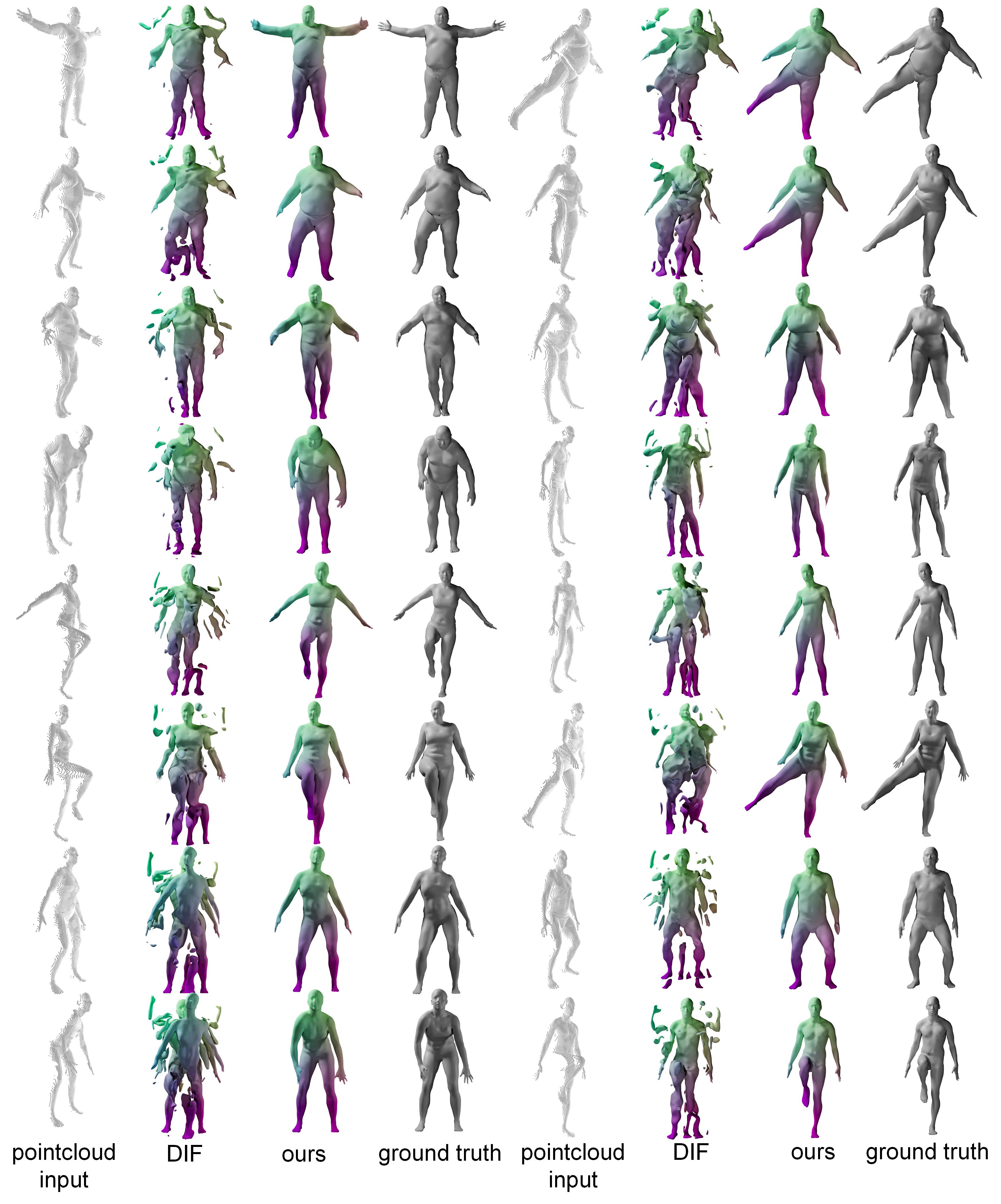}
    \caption{Reconstruction from partial observation of humans. We compare our method with DIF~\cite{Deng_2021_CVPR}. Our method reconstructs shapes with multiple poses and large deformations. The characteristics of each subject is represented well.}
    \label{fig:comppartialhuman}
\end{figure*}

\begin{figure*}[b]
    \centering
    \includegraphics[width=1.0\textwidth]{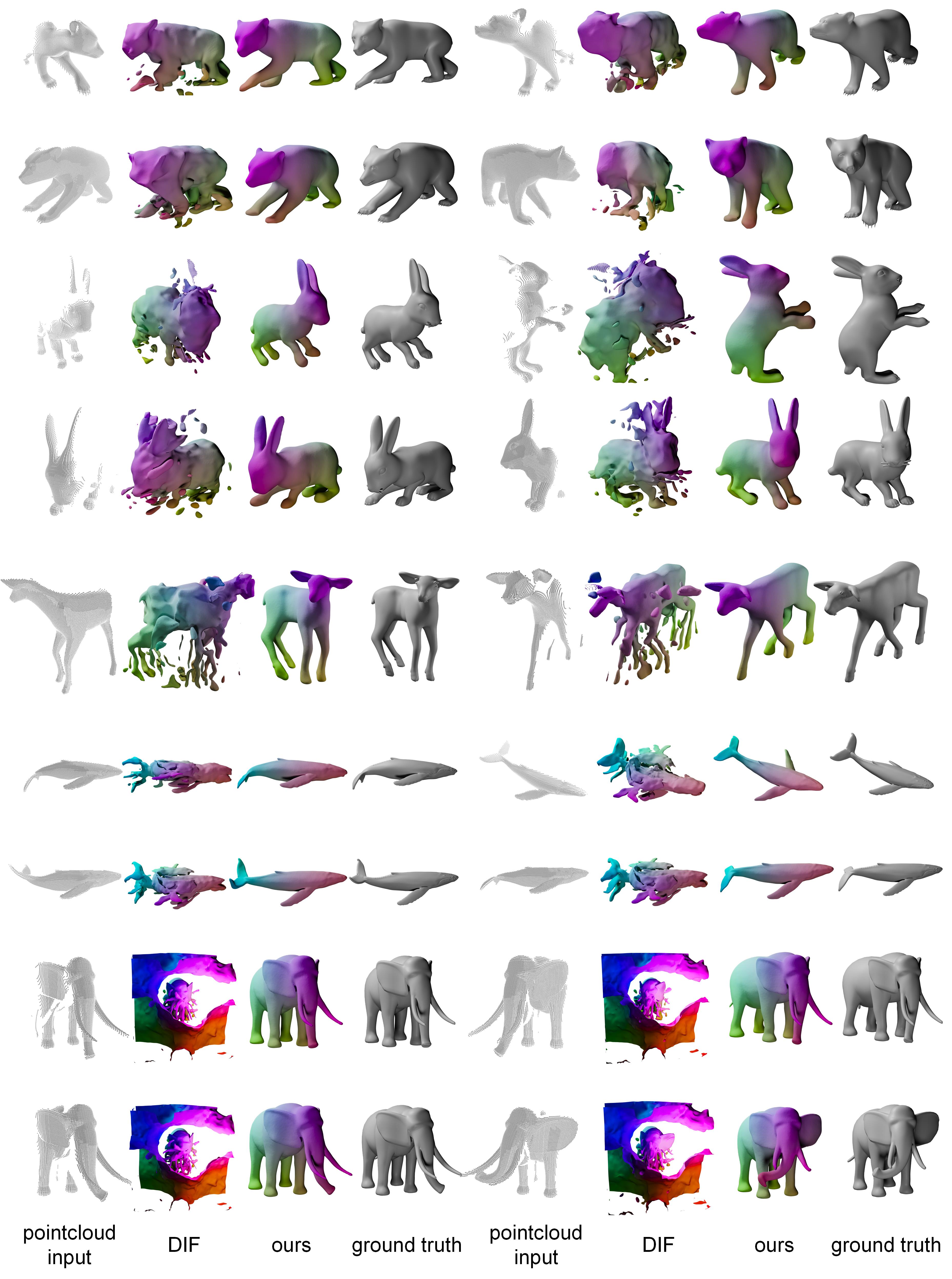}
    \caption{Reconstruction from partial observation of animals. We compare our method with DIF~\cite{Deng_2021_CVPR}. Our method reconstructs shapes with multiple poses and large deformations. The characteristics of each subject is represented well.}
    \label{fig:comppartialanimal}
\end{figure*}

We show several failure cases in Figure~\ref{fig:flyer} where floating components make Chamfer distance increase.
Although these shapes have fine human surface geometry, they have large Chamfer distance because of the floating components far from the body.

\section{Details on Rigid Constraint}
In this section, we will give details on the closed-form solution of piece-wise rigid constraint in Section 3.3 of the main paper.
Then, we will give theoretic analysis on our local rigid constraint, in which we elaborate on the relationship between the proposed constraint with implicit representation in Section 3.2 of the main paper and the traditional As-Rigid-As-Possible loss \cite{sorkine2017least} that was originally proposed for discrete mesh deformation.

\subsection{Closed-Form Solution of Piece-wise Rigid Constraint}
\label{sec:closeform}
We follow Sorkine-Hornun~\etal \cite{sorkine2017least} to give a closed-form solution of the minimal rigid transformation error of our piece-wise rigid
constraint $L_{pr}$ (Eq.~6 of our main paper). Detailed proof can be found in \cite{sorkine2017least}. In this section, we use the same notions as \cite{sorkine2017least} for easy understanding.

Denote $\mathcal{P}=\{\mathbf{p}_1,\mathbf{p}_2,...,\mathbf{p}_n\}$ and $\mathcal{Q}=\{\mathbf{q}_1,\mathbf{q}_2,...,\mathbf{q}_n\}$ to be corresponding points in $\mathbb{R}^d$. 
Therefore, the optimal rigid transformation $(\mathbf{R},\mathbf{t})$ between $\mathcal{P}$ and $\mathcal{Q}$ can be estimated by minimizing the following alignment error as
\begin{equation}
\begin{aligned}
L &= \min_{\mathbf{R},\mathbf{t}}F(\mathbf{R},\mathbf{t})\\
F &= \sum_{i=1}^n{w_i\Vert \mathbf{R}\mathbf{p}_i + \mathbf{t}- \mathbf{q}_i\Vert^2},
\end{aligned}
\label{eq:defF}
\end{equation}
where $\mathbf{R}\in SO(d)$ is rotation matrix and $\mathbf{t}\in \mathbb{R}^d$ is translation.

First, Sorkine-Hornun~\etal \cite{sorkine2017least} proved that the optimal translation $\mathbf{t}$ can be expressed as
\begin{equation}
    \mathbf{t}=\bar{\mathbf{q}}-\mathbf{R}\bar{\mathbf{p}},
\end{equation}
where $\bar{\mathbf{q}}$ and $\bar{\mathbf{p}}$ are the centroid of $\mathcal{Q}$ and $\mathcal{P}$ 
\begin{equation}
    \bar{\mathbf{p}}=\frac{\sum_{i=1}^n {w_i \mathbf{p}_i}}{\sum_{i=1}^n w_i}, \quad 
    \bar{\mathbf{q}}=\frac{\sum_{i=1}^n {w_i \mathbf{q}_i}}{\sum_{i=1}^n w_i}.
\label{eq:mean}
\end{equation}

Incorporate the optimal $\mathbf{t}$ into Eq.~\ref{eq:defF}, and then we get the loss function $F$ as
\begin{equation}
    F=\sum_{i=1}^n {w_i\Vert \mathbf{R}(\mathbf{p}_i-\bar{\mathbf{p}}) - (\mathbf{q}_i-\bar{\mathbf{q}}) \Vert^2}.
\end{equation}

Giving the definitions as follows
\begin{equation}
    \mathbf{x}_i:=\mathbf{p}_i-\bar{\mathbf{p}},
    \quad
    \mathbf{y}_i:=\mathbf{q}_i-\bar{\mathbf{q}},
\label{eq:centering}
\end{equation}
we can set the translation $\mathbf{t}$ to be zero, and then  focus on the estimation of $\mathbf{R}$ by optimizing the following equivalent loss function
\begin{equation}
    L=\min_{\mathbf{R}} {\sum_{i=1}^n {w_i\Vert \mathbf{R}\mathbf{x}_i-\mathbf{y}_i \Vert^2}}.
\label{eq:minR}
\end{equation}

Denote $\mathbf{W}=diag(w_1,...,w_n)$, $\mathbf{X}=[\mathbf{x}_1,\mathbf{x}_2,...,\mathbf{x}_n]$, $\mathbf{Y}=[\mathbf{y}_1,\mathbf{y}_2,...,\mathbf{y}_n]$. 
Then the loss function $L$ can be effectively calculated with the following closed-form solution
\begin{equation}
\begin{aligned}    &L=\sum_{i=1}^n{w_i (\Vert \mathbf{x}_i\Vert^2 + \Vert \mathbf{y}_i\Vert^2)-2 S_\sigma(\mathbf{X}\mathbf{W}\mathbf{Y}^T)}\\
    S_\sigma=&\begin{cases}
    \sigma_1+\sigma_2+...+\sigma_{d-1}+\sigma_d,\ if\ det(\mathbf{U}\mathbf{V}^T)=1\\
    \sigma_1+\sigma_2+...+\sigma_{d-1}-\sigma_d,\ if\ det(\mathbf{U}\mathbf{V}^T)=-1,
    \end{cases}
\end{aligned}
\label{eq:closedform}
\end{equation}
where $\mathbf{U}$, $\mathbf{V}$ are the left and the right singular matrices of $\mathbf{X}\mathbf{W}\mathbf{Y}^T$, and $\{\sigma_i\}$ are singular value of $\mathbf{X}\mathbf{W}\mathbf{Y}^T$ in descending order. Moreover, the gradient of $S_\sigma$ is a rotation matrix, which does not contain large value and enables stable learning process.

In order to calculate the 3D alignment error $L$, we only need several  efficient operations, such as solving  SVD of $3\times3$ square matrix $\mathbf{X}\mathbf{W}\mathbf{Y}^T$ and conducting point-wise additions and multiplications.

In our problem, $\mathcal{P}$ and $\mathcal{Q}$ are consisting of the points in the target space $\{\mathbf{p}\}$ and the correspondence $\{D_{i\rightarrow tmpl}(\mathbf{p})\}$ in the template space, respectively. 
Therefore, the piece-wise rigid loss $L_{pr}$ (Eq.~6 of our main paper) can be expressed in closed-form as
\begin{equation}
\begin{aligned}
    &\min_{\mathbf{R}_h,\mathbf{t}_h}\sum_{\mathbf{p}\in{S_i^0\cup S_i^-}}
    \mathbf{P}_{h}(\mathbf{p}) \Vert (\mathbf{R}_h\mathbf{p}+\mathbf{t}_h)
    -{D_{i \rightarrow tmpl}(\mathbf{p})}\Vert_2^2\\
    =&\sum_{\mathbf{p}\in{S_i^0\cup S_i^-}}\mathbf{P}_h(\mathbf{p})(\Vert \mathbf{x}\Vert_2+\Vert \mathbf{x}_{i\rightarrow tmpl}\Vert_2)-2S_\sigma(\mathbf{X}\mathbf{W}_h\mathbf{X}_{i\rightarrow tmpl}^T)
\end{aligned}
\label{eq:closedform2}
\end{equation}
where $\mathbf{x}$
and $\mathbf{x}_{i\rightarrow tmpl}$ are the points in the target space and their correspondence in template after removing their respective centroid as 
\begin{equation}
\begin{aligned}
    &\mathbf{x} = \mathbf{p}-\bar{\mathbf{p}}\\
    &\mathbf{x}_{i\rightarrow tmpl} = D_{i \rightarrow tmpl}(\mathbf{p})-\bar{D}_{i \rightarrow tmpl}(\mathbf{p}),\\
\end{aligned}
\end{equation}
and $\bar{\mathbf{p}}$ and $\bar{D}_{i \rightarrow tmpl}(\mathbf{p})$ are their respective centroid as Eq.~\ref{eq:mean}.
The $j$-th column of $\mathbf{X}\in \mathbb{R}^{3\times n}$ is a $\mathbf{x}$ derived from the $j$-th point $\mathbf{p}$,
each column of $\mathbf{X}_{i\rightarrow tmpl}\in \mathbb{R}^{3\times n}$ is the correspondence $\mathbf{x}_{i\rightarrow tmpl}$ of the $j$-th point $\mathbf{p}$, 
$\mathbf{P}_h(\mathbf{p})$ is the probability that point $\mathbf{p}$ belongs to $h$-th part, and $\mathbf{W}_h$ is a $n\times n$ diagonal matrix, its $j$-th diagonal element is the predicted part probability $\mathbf{P}_h$ of the $j$-th point $\mathbf{p}$.

\subsection{Analysis on Local Rigid Constraint}
\subsubsection{Analysis on Least Square Solution of Rotation}
\label{sec:leastrotation}
In this section, we give further analysis on the formulation of closest
rotation matrix of $J(D_{i\rightarrow tmpl})$.
With singular value decomposition (SVD), we get $J(D_{i\rightarrow tmpl})=\mathbf{U}\mathbf{\Sigma}\mathbf{V}^T$.
Given the properties of the determinant, we know that $det(J(D_{i\rightarrow tmpl}))=det(\mathbf{U})det(\mathbf{\Sigma})det(\mathbf{V}^T)$.
According to the definition of the singular value, the singular values of $J(D_{i\rightarrow tmpl})$ (\ie diagonal items of $\mathbf{\Sigma}$) are always positive.
Therefore, $det(J(D_{i\rightarrow tmpl}))$ has the same sign as $det(\mathbf{U})det(\mathbf{V}^T)$, \ie $det(\mathbf{U}\mathbf{V}^T)$.
When $det(J(D_{i\rightarrow tmpl}))<0$, its closest orthogonal matrix $\mathbf{U}\mathbf{V}^T$ has negative determinant.
However, a rotation matrix must have positive determinant.
To this end, previous method \cite{umeyama1991least} figured out the closet rotation that has positive determinant. $\mathbf{R} = \mathbf{U}\mathbf{S}\mathbf{V}^T$ ($\mathbf{R}$ have positive determinant) of $J(D_{i\rightarrow tmpl})$ with a diagonal matrix $\mathbf{S}=diag(1,1,det(\mathbf{U}\mathbf{V}^T))$.

\subsubsection{Equivalence between Local Rigid Constraint and ARAP}
\label{sec:arap}
In this section, we will prove that 
our implicit local rigid constraint is equivalent to traditional As-Rigid-As-Possible (ARAP) loss in infinite small scope. ARAP loss \cite{sorkine2007rigid} is generally defined on discrete representations such as mesh, while we find that with the closed form of alignment error Eq.~\ref{eq:closedform} ARAP loss can be extended to continuous implicit representation for infinite small scope.

According to Sorkine \etal \cite{sorkine2007rigid}, ARAP loss on mesh is defined as
\begin{equation}
    E=\min_{\mathbf{R}}\sum_{j\in\mathcal{N}(i)}w_{ij}\Vert(\mathbf{p}'_i-\mathbf{p}'_j)-\mathbf{R}_i(\mathbf{p}_i-\mathbf{p}_j)\Vert^2.
\label{eq:arap}
\end{equation}

In our method, we represent the shape as implicit field instead of mesh in original ARAP \cite{sorkine2007rigid}, so there is not explicit adjacency relation for our shape representation. It is barely addressed and highly challenging to constrain ARAP in the continuous implicit shape representation.

For a sampling point $\mathbf{p}$, we assume the adjacent points of $\mathbf{p}$ are uniformly distributed on the surface of a sphere centered at $\mathbf{p}$, which can be formulated as $\mathbf{p}+\mathbf{\omega}s$, where $\mathbf{\omega}$ is an arbitrary unit vector.
For simplicity, we consider $w_{ij}$ as 1.

Considering the adjacent points within the infinitely-small volume, we denote adjacent points as 
\begin{equation}
    \mathbf{p}+\mathbf{\omega}ds,
\end{equation}
where $ds$ is infinitely small length.

Denote the mapping from $\mathbf{p}$ to $\mathbf{p}'$ as $D(\mathbf{p})$ and $\frac{\partial \mathbf{p}'}{\partial\mathbf{p}^T}$, \ie $J(D_{i\rightarrow tmpl})(\mathbf{p})$, as $J$.
In our case, $\mathbf{p}$ is in the target shape space and $\mathbf{p}'$ is in the template shape space. Then we have the following equation by Taylor expansion
\begin{equation}
    D(\mathbf{p}+\mathbf{\omega}ds)= D(\mathbf{p})+J\mathbf{\omega}ds + o(ds).
\label{eq:mapping}
\end{equation}

According to Eq.~\ref{eq:closedform}, we can also get a closed-form solution for ARAP loss in Eq.~\ref{eq:arap}. Because $\mathbf{\omega}$ is evenly distributed on the sphere, $\sum_\mathbf{\omega} \mathbf{\omega}=\mathbf{0}$, the centroid of $\mathbf{p}+\mathbf{\omega}ds$ is $\mathbf{p}$ and the centroid of $D(\mathbf{p})+J\mathbf{\omega}ds$ is $D(\mathbf{p})$. After incorporating $\mathbf{p}+\mathbf{\omega}ds$ and $D(\mathbf{p})+J\mathbf{\omega}ds$ and their centroids into Eq.~\ref{eq:centering}, we get $\mathbf{x}_i=\mathbf{\omega}ds$ and $\mathbf{y}_i=J\mathbf{\omega}ds+o(ds)$, and incorporate $\mathbf{x}_i$ and $\mathbf{y}_i$ into Eq.~\ref{eq:closedform}, then we can get the following equation by ignoring the infinitesimal of higher order
\begin{equation}
\begin{aligned}
    E=&\sum_{\mathbf{\omega}} \Vert \mathbf{\omega}ds\Vert^2 + \Vert J\mathbf{\omega}ds\Vert^2-2S_\sigma(\sum_{\mathbf{\omega}}{\mathbf{\omega}\mathbf{\omega}^TJ^T}ds^2)\\
    =&ds^2\sum_{\mathbf{\omega}} (\Vert \mathbf{\omega}\Vert^2 + \Vert J\mathbf{\omega}\Vert^2)-2ds^2S_\sigma(\sum_{\mathbf{\omega}}{\mathbf{\omega}\mathbf{\omega}^TJ^T}).
    \end{aligned}
\end{equation}
Since $\omega$ is uniformly distributed, we use integration instead of summation.
\begin{equation}
\begin{aligned}
    E=ds^2(\int_{\mathcal{S}^2}\Vert \mathbf{\omega}\Vert^2 d\mathbf{\omega} + &\int_{\mathcal{S}^2}\Vert J\mathbf{\omega}\Vert^2 d\mathbf{\omega}\\
    -2S_\sigma(&\int_{\mathcal{S}^2}{\mathbf{\omega}\mathbf{\omega}^TJ^T}d\mathbf{\omega})),
\end{aligned}
\label{eq:totalform}
\end{equation}
where $S^2$ represents the surface of a unit sphere embedded in the 3-dimensional space, and each term will be analyzed  in the following part.

Since $\Vert\mathbf{\omega}\Vert^2=1$, the first term can be easily calculated as the area of the sphere, \ie $4\pi$.

Then the second term can be simplified as
\begin{equation}
    \begin{aligned}
        &\quad\int_{\mathcal{S}^2}\Vert J\mathbf{\omega}\Vert^2 d\mathbf{\omega}
        =\int_{\mathcal{S}^2}\mathbf{\omega}^TJ^TJ\mathbf{\omega}d\mathbf{\omega}\\
        &=\int_{\mathcal{S}^2}tr(\mathbf{\omega}^TJ^TJ\mathbf{\omega})d\mathbf{\omega}
        =\int_{\mathcal{S}^2}tr(\mathbf{\omega}\mathbf{\omega}^TJ^TJ)d\mathbf{\omega}\\
        &=tr\Bigg(\int_{\mathcal{S}^2}J^TJ\mathbf{\omega}\mathbf{\omega}^Td\mathbf{\omega}\Bigg)
        =tr\Bigg(J^TJ\int_{\mathcal{S}^2}\mathbf{\omega}\mathbf{\omega}^Td\mathbf{\omega}\Bigg).
    \end{aligned}
    \label{eq:secondtermfull}
\end{equation}
To solve the above function, we need to know the result of $\int_{\mathcal{S}^2}\mathbf{\omega}\mathbf{\omega}^Td\mathbf{\omega}$.
We use spherical coordinates to calculate the integration.
$\mathbf{\omega}=(sin\theta cos\phi,sin\theta sin\phi,cos\theta)^T=(sin\theta cos\phi,sin\theta sin\phi,0)^T+(0,0,cos\theta)^T$
\begin{equation}
    \begin{aligned}
    \int_{\mathcal{S}^2}\mathbf{\omega}\mathbf{\omega}^Td\mathbf{\omega}=\\
    \int_0^\pi\int_0^{2\pi}sin\theta
    \Bigg(&
    \begin{pmatrix}
    0\\
    0\\
    cos\theta
    \end{pmatrix}+
    \begin{pmatrix}
    sin\theta cos\phi\\
    sin\theta sin\phi\\
    0
    \end{pmatrix}
    \Bigg)\\
    &\Bigg(
    \begin{pmatrix}
    0\\
    0\\
    cos\theta
    \end{pmatrix}+
    \begin{pmatrix}
    sin\theta cos\phi\\
    sin\theta sin\phi\\
    0
    \end{pmatrix}
    \Bigg)^Td\phi d\theta\\
    \end{aligned}
\end{equation}
Consider $\int_0^{2\pi} sin\phi\ d\phi = 0$ and $\int_0^{2\pi} cos\phi\ d\phi = 0$:
\begin{equation}
    \begin{aligned}
    =\int_0^\pi\int_0^{2\pi}sin\theta
    &\Bigg(\begin{pmatrix}
    0\\
    0\\
    cos\theta
    \end{pmatrix}
    \begin{pmatrix}
    0\\
    0\\
    cos\theta
    \end{pmatrix}^T\\
    &+\begin{pmatrix}
    sin\theta cos\phi\\
    sin\theta sin\phi\\
    0
    \end{pmatrix}
    \begin{pmatrix}
    sin\theta cos\phi\\
    sin\theta sin\phi\\
    0
    \end{pmatrix}^T\Bigg)d\phi d\theta\\
    =\int_0^\pi\int_0^{2\pi}sin\theta
    &\begin{pmatrix}
    sin^2\theta cos^2\phi & \frac{sin^2\theta sin2\phi}{2} & 0\\
    \frac{sin^2\theta sin2\phi}{2} & sin^2\theta sin^2\phi & 0\\
    0 & 0 & cos^2\theta
    \end{pmatrix}d\phi d\theta
    \end{aligned}
\end{equation}
Consider the periodicity of $\int_0^{2\pi} sin2\phi\ d\phi = 0$:
\begin{equation}
    \begin{aligned}
    =&\int_0^\pi\int_0^{2\pi}sin\theta
    \begin{pmatrix}
    sin^2\theta cos^2\phi & 0 & 0\\
    0 & sin^2\theta sin^2\phi & 0\\
    0 & 0 & cos^2\theta
    \end{pmatrix}d\phi d\theta\\
    =&\begin{pmatrix}
    \frac{4\pi}{3} & 0 & 0\\
    0 & \frac{4\pi}{3} & 0\\
    0& 0 & \frac{4\pi}{3}
    \end{pmatrix}.
    \end{aligned}
\end{equation}

Denote singular value decomposition (SVD) of $J$ to be $J=\mathbf{U}\mathbf{\Sigma}\mathbf{V}^T$.
Then Eq.~\ref{eq:secondtermfull} can be calculated as
\begin{equation}
    \begin{aligned}
        &tr(J^TJ\int_{\mathcal{S}^2}\mathbf{\omega}\mathbf{\omega}^Td\mathbf{\omega})=tr(J^TJ\mathbf{I}\frac{4\pi}{3})\\
    =&\frac{4\pi}{3}tr(J^TJ)=\frac{4\pi}{3}tr(\mathbf{V}\mathbf{\Sigma}\mathbf{U}^T\mathbf{U}\mathbf{\Sigma}\mathbf{V}^T)\\
    =&\frac{4\pi}{3}tr(\mathbf{V}\mathbf{\Sigma}
    \mathbf{\Sigma}\mathbf{V}^T)=\frac{4\pi}{3}tr(\mathbf{V}^T\mathbf{V}\mathbf{\Sigma}
    \mathbf{\Sigma})\\
    =&\frac{4\pi}{3}tr(\mathbf{\Sigma}^2)=\frac{4\pi}{3}(\sigma_1^2+\sigma_2^2+\sigma_3^2),
    \end{aligned}
\end{equation}
where $\sigma_1$, $\sigma_2$, $\sigma_3$ are singular values in descending order.

With the above result of $\int_{\mathcal{S}^2}\mathbf{\omega}\mathbf{\omega}^Td\mathbf{\omega}$, the third term of Eq.~\ref{eq:totalform} can be calculated as
\begin{equation}
\begin{aligned}
    &S_\sigma\Bigg(\int_{\mathcal{S}^2}{\mathbf{\omega}\mathbf{\omega}^TJ^T}d\mathbf{\omega}\Bigg)\\
    =&S_\sigma\Bigg(\int_{\mathcal{S}^2}({\mathbf{\omega}\mathbf{\omega}^TJ^T})^Td\mathbf{\omega}\Bigg)\\
    =&S_\sigma\Bigg(J\int_{\mathcal{S}^2}{\mathbf{\omega}\mathbf{\omega}^T}d\mathbf{\omega}\Bigg)=S_\sigma\Bigg(\frac{4\pi}{3}J\Bigg)\\
    =&\frac{4\pi}{3}(\sigma_1+\sigma_2+det(\mathbf{U}\mathbf{V^T})\sigma_3).
\end{aligned}
\end{equation}

We can simplify Eq.~\ref{eq:totalform} with the above results of its three items as
\begin{equation}
\begin{aligned}
    E&=\frac{4\pi}{3}ds^2(3+\sigma_1^2+\sigma_2^2+\sigma_3^2-2(\sigma_1+\sigma_2+det(\mathbf{U}\mathbf{V^T})\sigma_3))\\
    &=\frac{4\pi}{3}ds^2((\sigma_1-1)^2+(\sigma_2-1)^2)+(\sigma_3-det(\mathbf{U}\mathbf{V^T}))^2).
\end{aligned}
\label{eq:arap2}
\end{equation}


Our $L_{arap}$ in Section 3.2 of our main paper has the following formulation
\begin{equation}
\begin{aligned}
&L_{arap}=smoothL1(\sigma_1,1) + smoothL1(\sigma_2,1)\\
& +smoothL1(\sigma_3,det(\mathbf{U}\mathbf{V}^T))
\end{aligned}
\end{equation}
If $\sigma_1$, $\sigma_2$ and $\sigma_3$ are close to 1 and $\mathbf{U}\mathbf{V}^T = 1$, the smoothL1 loss becomes L2 loss. 
By ignoring the scale term, $L_{arap}$ has the same form as Eq.~\ref{eq:arap2}.

Therefore, our implicit local rigid constraint is equivalent to traditional As-Rigid-As-Possible (ARAP) loss in infinite small scope.


{\small
\bibliographystyle{ieee_fullname}
\bibliography{egbib}
}

\end{document}